%% file: main.tex
\documentclass{article} 
\usepackage{iclr2027/iclr2027_conference,times}

\input{iclr2027/math_commands.tex}

\usepackage[hidelinks]{hyperref}
\usepackage{url}
\usepackage{graphicx}
\usepackage{wrapfig}
\usepackage{capt-of}
\usepackage{booktabs}
\usepackage{array}
\usepackage{multirow}
\usepackage{makecell}
\usepackage[table]{xcolor}
\usepackage{bbm}
\usepackage{enumitem}
\usepackage{tcolorbox}

\newcommand{\modelname}{UserLM-2.0}
\newcommand{\modelnamesft}{UserLM-2.0-SFT}
\newlength{\tablegroupsep}
\newcolumntype{/}{!{\color{white}\vline width \tablegroupsep}}

\title{Towards User Simulators That Stay on Task and Still Sound Natural}

\author{%
Mustafa Omer Gul\textsuperscript{1}\thanks{Work done during an internship at Microsoft Research.}\quad
Philippe Laban\textsuperscript{2}\quad
Tanya Goyal\textsuperscript{2}\quad
Jennifer Neville\textsuperscript{2}\\
{\normalfont\textsuperscript{1}Cornell University \quad \textsuperscript{2}Microsoft Research}\\
{\normalfont\small\texttt{momergul@cs.cornell.edu ; \{plaban,tanyagoyal,jenneville\}@microsoft.com}}
}

\iclrfinalcopy 
\begin{document}

\maketitle
\lhead{}
\vspace{-26pt}

\input{sections/00_abstract}
\input{sections/10_intro}
\input{sections/20_training}

\input{sections/30_main_evaluation}
\input{sections/40_analysis}
\input{sections/50_related_work}
\input{sections/60_conclusion}

\bibliography{iclr2027/iclr2027_conference}
\bibliographystyle{iclr2027/iclr2027_conference}

\input{sections/appendix}

\end{document}

%% file: iclr2027/math_commands.tex
\usepackage{amsmath,amsfonts,bm}

\def\eqref#1{equation~\ref{#1}}

\def\1{\bm{1}}

\DeclareMathAlphabet{\mathsfit}{\encodingdefault}{\sfdefault}{m}{sl}
\SetMathAlphabet{\mathsfit}{bold}{\encodingdefault}{\sfdefault}{bx}{n}



%% file: sections/00_abstract.tex
\begin{abstract}
Users increasingly rely on AI systems for extended, multi-turn tasks, yet evidence indicates that LLM-based system performance degrades over prolonged interaction. To evaluate and improve AI assistants in such settings, we need user simulators that faithfully follow complex user intents while exhibiting realistic conversational behavior. Existing approaches manifest a trade-off: prompted assistants can follow complex user intents but lack behavioral fidelity, while specialized user simulators better approximate conversational behavior but struggle to remain faithful to complex intents. In this paper, we propose a reinforcement learning (RL) recipe that jointly rewards task intent adherence and behavioral naturalness. A crucial contribution of the method is operationalizing naturalness with a lightweight discriminator trained in an alternating, adversarial loop with the user simulator. We release the resulting model, \textbf{\modelname{}}, which produces valid simulated conversations (i.e., meeting both task-faithfulness and naturalness criteria) 46.2\% of the time, over three times the rate of the strongest prior user simulator (14.1\%).
\end{abstract}
\vspace{-4pt}
\begingroup
\setlength{\intextsep}{1pt}
\begin{figure}[!ht]
    \centering
    \setlength{\abovecaptionskip}{5pt}
    {\small\ttfamily\href{https://huggingface.co/microsoft/UserLM-2.0}{\raisebox{\dimexpr0.5\fontcharht\font`X-0.5\height\relax}{\includegraphics[height=1em]{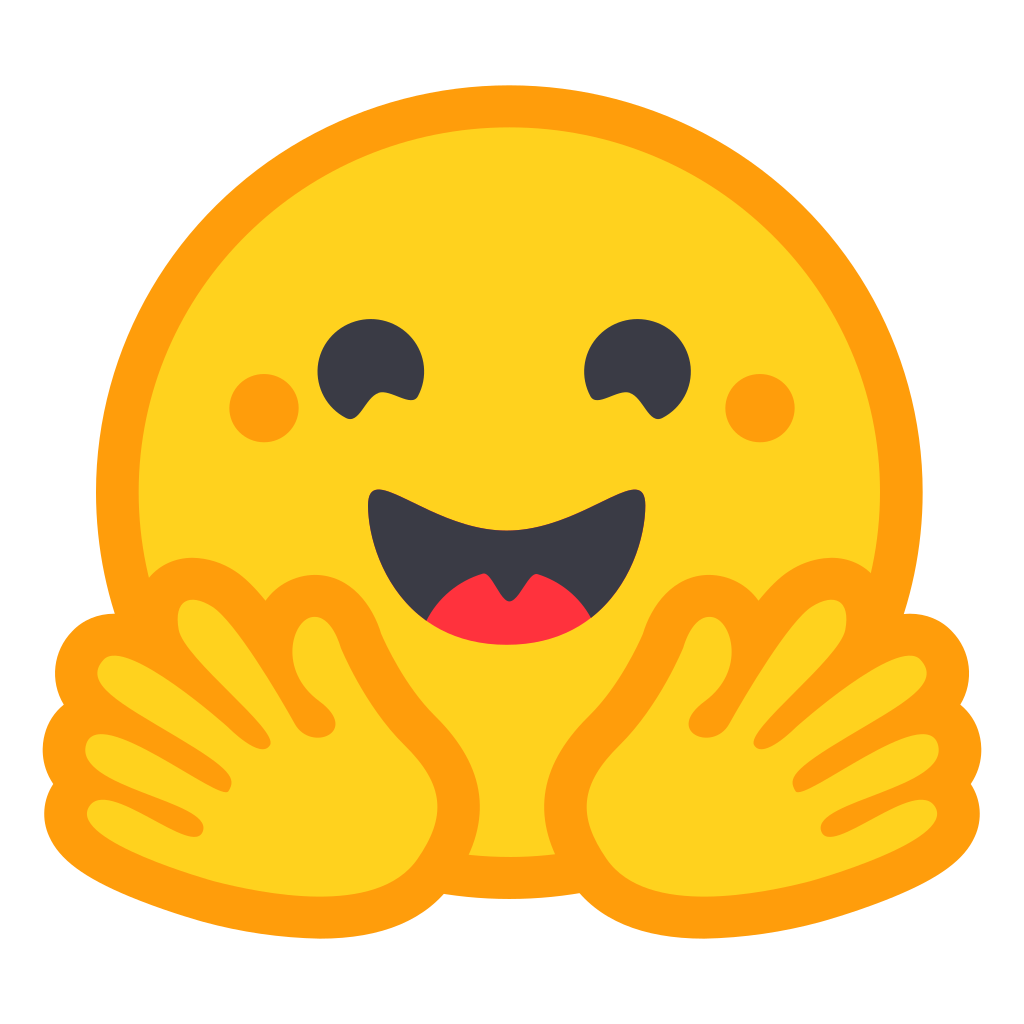}}~\textcolor{blue}{microsoft/UserLM-2.0}}\par}
    \vspace{3pt}
    \includegraphics[trim={0 50 0 0}, clip, width=\textwidth]{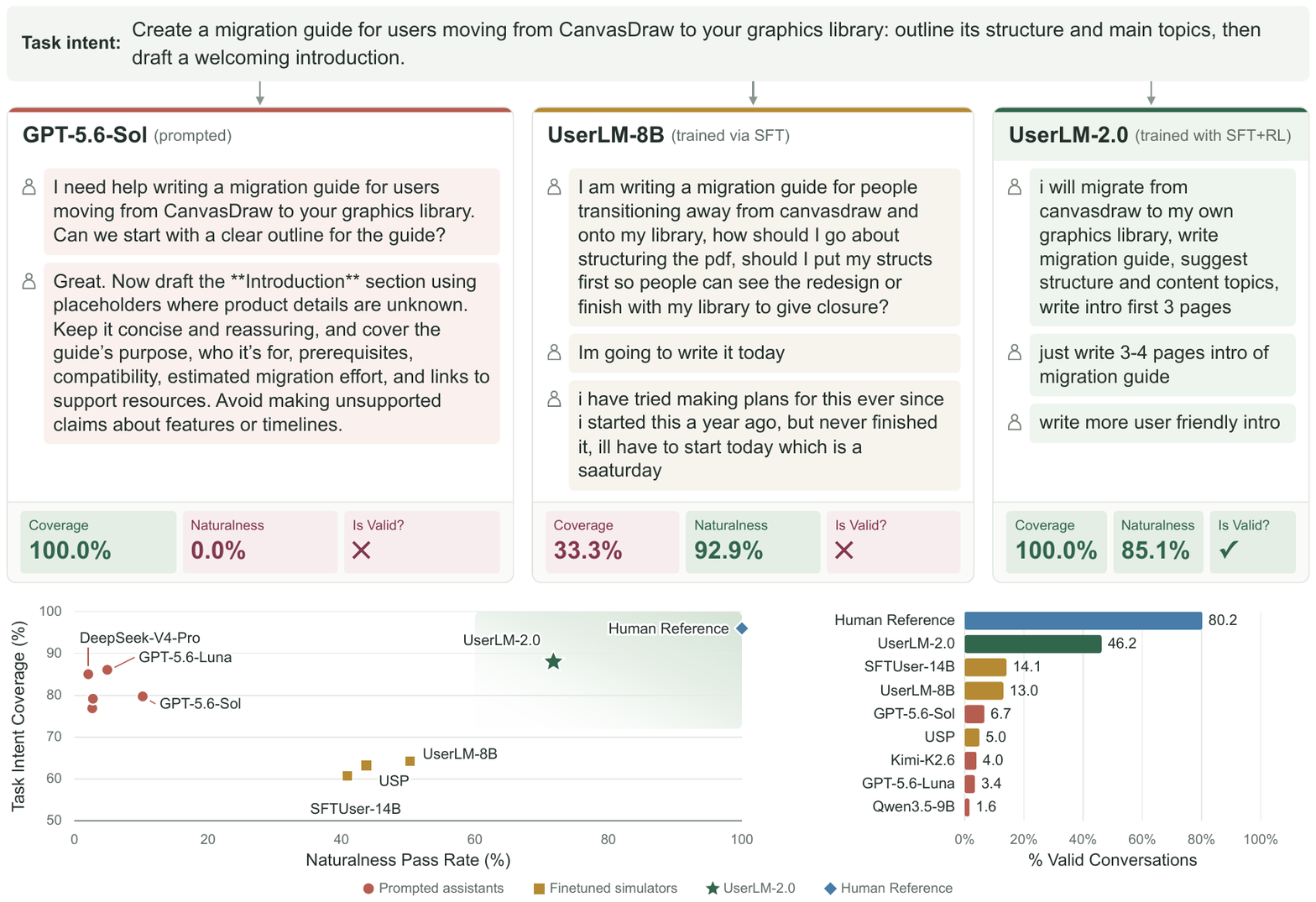}
    \caption{We train and release \modelname{}, a user simulator for User-AI interaction optimized via RL to precisely enact user intents while exhibiting natural conversational behavior. In comparison, assistants such as GPT-5.6-Sol follow intents, but sound unnatural, while SFT-trained simulators like UserLM-8B~\citep{naous2026flipping} generate natural, but meandering conversations.}
    \label{fig:teaser}
\end{figure}
\endgroup

%% file: sections/10_intro.tex
\section{Introduction}
Users increasingly expect AI assistants and agents to help them accomplish complex tasks through sustained, multi-turn interaction, rather than provide isolated answers to individual queries~\citep{handa2025economictasksperformedai,chatterji2025people}. Yet prior work has shown that current LLM-based systems severely degrade in performance in such interactive settings~\citep{laban2025llmslostmultiturnconversation,tack2026llmslostevolvinguser}.

Understanding and addressing this gap requires evaluations that capture how real users communicate their goals and respond to assistants over multiple turns~\citep{chang-etal-2025-chatbench}. A crucial component of interactive benchmarks is a \textit{user simulator}, which takes as input a high-level \textit{task intent} (such as the one shown in Figure~\ref{fig:teaser}) and generates plausible user utterances through iterative interaction with an assistant~\citep{yao2024taubenchbenchmarktoolagentuserinteraction,lu2025toolsandboxstatefulconversationalinteractive,huang2025crmarenaproholisticassessmentllm}.

We focus on two essential properties of user simulators: task fidelity and behavioral naturalness. Task fidelity requires the simulator to convey the information specified in the task intent over the course of the conversation, without omitting or contradicting the specification. Behavioral naturalness requires the simulator to resemble real users in both how it phrases individual messages and how it conducts the interaction across turns. A simulator that omits task requirements can make assistant models appear to fail when the simulator is at fault, making benchmarking unreliable~\citep{mehri2026goalalignmentllmbaseduser,zhou2026mind}, whereas a simulator that does not adequately capture the interactive behavior of real users may yield conclusions that do not transfer to real interactions~\citep{zhou2026mind,seshadri2026lostsimulationllmsimulatedusers}.

We show experimentally that current approaches favor one requirement at the expense of the other. Prompting LLM assistants to act as users is the most common method for simulating users~\citep{yao2024taubenchbenchmarktoolagentuserinteraction,lu2025toolsandboxstatefulconversationalinteractive,huang2025crmarenaproholisticassessmentllm}. While prompted assistants can stay on task, they lack behavioral naturalness~\citep{naous2026flipping,mehri2026measuring,zhou2026mind}. This is illustrated in the example in Figure~\ref{fig:teaser}, where GPT-5.6-Sol (a strong assistant LLM) accurately enacts the task intent but expresses it through unusually polished and exhaustive messages, unlike the conversational patterns in real user-AI exchanges.

An alternative is to train user simulators by ``flipping the dialogue'': supervised finetuning an LLM on user utterances from a corpus of user-assistant interactions~\citep{naous2026flipping,suh2026quantifyingutilityusersimulators}. As reported in prior work, such finetuned user simulators exhibit realistic interaction behavior but often struggle to precisely enact the task intent they are given~\citep{naous2026flipping,zhou2026mind}. In the example in Figure~\ref{fig:teaser}, the SFT-trained UserLM-8B~\citep{naous2026flipping} ends the conversation prematurely, before conveying all of the specified task requirements to the assistant.

We propose a post-training recipe that improves task fidelity while preserving behavioral naturalness in user simulators. We first finetune an LLM on user turns from a corpus of user-assistant interactions, then post-train it using multi-turn reinforcement learning.

Our RL training objective jointly rewards task fidelity and behavioral naturalness. Rewards related to task fidelity use LLM-based judges~\citep{zheng2023judgingllmasajudgemtbenchchatbot}, for example by evaluating the extent to which a simulated conversation has conveyed the information specified in the task intent. We find that LLM-as-a-judge for behavioral naturalness fails to reliably encourage realistic user behavior. Instead, drawing on adversarial training~\citep{NIPS2014_f033ed80} and its application to dialogue generation~\citep{li-etal-2017-adversarial}, we propose a discriminator-based reward for behavioral naturalness. Specifically, we continually retrain a discriminator to distinguish real user conversations from conversations generated by the user simulator being trained. This choice defines the naturalness reward in terms of how indistinguishable a given simulated conversation is from real user conversations. We update the discriminator as the simulator evolves to keep the naturalness reward responsive to its changing outputs and reduce opportunities to exploit a fixed evaluator~\citep{gao2022scalinglawsrewardmodel}.

We train, evaluate, and release the resulting artifact, \modelname{}. Among the simulators we evaluated, \modelname{} lies on the Pareto frontier for task fidelity and behavioral naturalness. We count a simulated conversation as valid if it adheres to our quality and termination criteria, achieves at least 80\% coverage of the task intent \textit{and} passes our naturalness criterion, assessed using three discriminators including Pangram~\citep{glickenhaus2026pangram}. By this definition, \modelname{} yields valid conversations for 46.2\% of task intents, more than three times the 14.1\% rate of the strongest prior user simulator (SFTUser-14B). This gap grows even further to a factor of 12 at higher task complexities, highlighting the effectiveness of our method.

%% file: sections/20_training.tex
\section{Method}

Our goal is to train a user simulator that takes a high-level task intent as input, and simulates realistic user utterances in a multi-turn user-assistant conversation. As previously discussed, a good user simulator must (1) generate natural utterances, and (2) obey the task intents closely.

\subsection{Problem Definition}\label{sec:problem_definition}

Each conversation is a multi-turn interaction where a user simulator $U$ interacts with an assistant model $A$ to satisfy an input intent $c$. This intent $c$ captures the high-level goal of the user without detailing behavior on a turn-by-turn basis. Given $c$, we sample conversations 
$\tau = (c, u_1, a_1, \ldots, u_{n-1}, a_{n-1}, u_n)$, where $u_i$ and $a_i$ represent the sampled $i$-th user and assistant utterances respectively. A conversation terminates either when the maximum number of user turns $N$ is reached or when the user outputs an $\texttt{<|endconversation|>}$ tag.

Prior work~\citep{wang-etal-2025-know, naous2026flipping} trains user simulators via SFT on a large corpus of user-assistant interactions. 
These exhibit the desired naturalness behavior, but are extremely poor at faithfully enacting task intents. 
We propose treating such SFT models as initializations rather than the final user simulator. We train our own SFT model, \modelnamesft{}, and design an RL-based post-training pipeline that improves its task fidelity without sacrificing naturalness.

We perform multi-turn reinforcement learning using the GRPO algorithm~\citep{shao2024deepseekmath}. Our reward balances the above  two goals of user simulation, task fidelity and behavioral naturalness. 
Although these are orthogonal, models can easily reward hack one at the cost of the other, such as by unnaturally including all information in the first utterance for perfect task coverage. The combined reward composition safeguards against such behaviors.

\subsection{Reward Design For Reinforcement Learning}
\label{sec:reward-design}
We first describe the individual reward components and then our formula for combining them:

\paragraph{Naturalness:} The naturalness reward should score how closely the user utterances of simulated conversation resemble the way real users interact with AI assistants in a target corpus. 
Let $P_\text{disc}$ be a discriminator that can distinguish between real and simulated user utterances. Our naturalness reward measures how well the user simulator fools such a discriminator based on $P_\text{disc}( y=\text{real} \mid \hat{u_1}...\hat{u_k}, c )$, where $\hat{u_1}...\hat{u_k}$ is the multi-turn output from the user simulator and $c$ is the associated task intent. For training, we set our naturalness reward to be binary; it is $1$ if this discriminator score is above a threshold ($0.5$ in our experiments) and $0$ otherwise.

Broadly, we have two options for discriminators: LLM judges or specially trained discriminators. LLM judges are ill-suited for this task. Naturalness captures multiple implicit dimensions such as the surface-form of the utterance, pace of the conversation, coherence of follow-up utterances, etc., all of which might be impossible to extract from the target corpus and enumerate as instructions. 
In fact, prior work~\citep{russell2025people} has shown that prompted LLM judges are ineffective at distinguishing between real user and AI text.\footnote{Note that AI detection of assistant turns is an  analogous task to user naturalness. There, the goal is to score the ``naturalness'' of AI assistants' text to distinguish it from human-written text. We highlight that the best AI detectors in industry use trained discriminators instead of LLM judges~\citep{glickenhaus2026pangram}.} In our work, we train lightweight discriminators on top of ModernBert-Large~\citep{warner-etal-2025-smarter} to distinguish between real and simulated user utterances. This allows us to both limit the costs and latency during RL and specialize to the user behaviors in the target corpus. We include ablations to empirically demonstrate the issues with using LLM judges and static discriminators in Section \ref{sec:analysis_naturalness_formulation}.

\paragraph{Coverage:} This reward component quantifies how completely the user simulator follows a task intent. We define coverage as the proportion of the sub-goals of the task intent that are included in the user utterances generated by the simulator. This is instantiated with an LLM judge.

We highlight that coverage can be an easy metric to reward hack for user simulators. For example, given a detailed task intent with fully specified sub-goals, a user simulator can maximize coverage by simply paraphrasing the sub-goals. This is not ideal as it disincentivizes diverse simulations for the same task intent. Therefore, we introduce an information asymmetry between the task intent provided to the user simulator as input ($c$) and the one used to compute the coverage reward by the LLM judge ($c^*$), where $c$ is a more abstract version of the fully-specified task intent $c^*$. This ensures the model receives poor coverage when simply paraphrasing, and encourages exploring more diverse behaviors.

In addition to coverage, we include two other reward components for task fidelity. These target proper formatting of the user utterances and elimination of trivial failure modes of the SFT models. 

\paragraph{Quality:}
Our quality reward is a binary score that assesses the baseline quality of a simulated conversation. It targets three common error modes of the SFT models: (1) verbatim repetition of task intent, (2) verbatim repetition of past user turns, and (3) switching roles from user to assistant mid-output. We first check the initial two conditions programmatically and otherwise use an LLM judge. A simulated conversation is given a reward $1$ if all three conditions are met, and $0$ otherwise.

\paragraph{Termination:} The goal of the termination reward is to ensure that user simulators properly terminate conversations  when all sub-goals in $c^*$ are met and penalize early termination. We give a positive reward $\lambda$ for including the termination tag \texttt{<|endconversation|>} in the last utterance if all sub-goals have been met in the earlier utterances and a negative reward $-\lambda$ if the conversation is terminated without satisfying all sub-goals. Note that this ties the termination reward to coverage. In some cases, the user simulator fails to terminate the conversation at all, i.e. does not generate the \texttt{<|endconversation|>} tag; in these cases, we give it a $0$ termination reward.

\paragraph{Final Reward:} We combine all four reward components into a overall reward score for a simulated conversation using the following formula:
$ r_\text{overall} = r_\text{quality} \cdot \left( r_\text{coverage} + r_\text{naturalness} + r_\text{termination} \right)$. Note that the overall reward is $0$ if the quality reward is $0$. Therefore, the basic quality checks serve as a necessary condition for any successful simulation.

\subsection{GAN-Style Training with Dynamic Discriminators} 
One limitation of trained naturalness discriminators, especially lightweight ones with fewer parameters, is that a larger user simulator trained against it will quickly learn to reward hack. It will quickly converge on a narrow set of behaviors that reliably fools the specific discriminator but does not learn broader behavioral naturalness. Our initial experiments validated this intuition -- user simulators trained against a static discriminator maximized the naturalness reward within 75 steps by greatly increasing average utterance length while reducing conversation turns. 

To address this, we propose a GAN-style training recipe that periodically re-trains the discriminator every k steps. This improves discriminator quality by exposing it to synthetic conversations with distinct behavioral patterns and further helps detect and discourage reward hacking behavior.

\section{Experimental Setup}\label{sec:experimental_setup} 
\subsection{Dataset Construction}\label{sec:dataset_construction} 

\paragraph{Dataset Processing.} We run our experiments on the non-toxic subset of WildChat-4.8M~\citep{zhao2024wildchat}.  
We build on top of the dataset processing scheme of \citet{naous2026flipping}, additionally removing interactions with only one user turn as our goal is to simulate multi-turn interactions. This reduces the WildChat-4.8M dataset size from 3,199,860 to 199,249 interactions.
We also follow \citet{naous2026flipping} in how we construct train, validation and test sets. 
We hold out 10,000 and 1,000 conversations from the train and validation set for RL training, 3,200 and 1,000 conversations from the train and validation set for discriminator training, and use the rest for SFT. Refer to Appendix \ref{sec:appendix_dataset_construction} for more details on dataset construction.

\paragraph{Intent and Subgoal Extraction.} We need to  extract both intents (given as input for simulation) and fine-grained sub-goals (used for LLM-based coverage reward). We use Qwen3-30B-A3B-Instruct-2507~\citep{yang2025qwen3technicalreport} to extract intents for all 199,249 conversations. For sub-goals, we few-shot prompt the best available GPT models without reasoning. Details are in Appendix \ref{sec:appendix_dataset_construction}. 

\paragraph{Test dataset.} We report all results on a 2000 intent test set, constructed such that the number of intents per sub-goal count (ranging from 1 to 7) is balanced. For each user simulator, we sample four conversations per intent  
with Qwen3.5-9B as the assistant.

\subsection{Training Details}\label{sec:training_details} 
\paragraph{User Simulator Training.} We use Llama3.1-8B-Instruct~\citep{grattafiori2024llama} as the backbone for \modelname{}. We replicate the SFT training of ~\citet{naous2026flipping} and select the checkpoint with the best validation loss for subsequent RL. This checkpoint is \modelnamesft{}. For RL, we simulate conversations with a maximum turn count of $N=5$, and use Qwen3.5-9B without reasoning~\citep{qwen35blog} as the assistant model. We set $\lambda=0.2$ for the termination reward, batch size to $64$, group size to $8$ and KL hyperparameter $\beta=0.001$. We measure model performance on the validation set every $25$ steps for checkpoint selection. See Appendix \ref{sec:appendix_training_details} for further training details.

\paragraph{Naturalness Discriminator Training.} We use the 395M ModernBERT-Large~\citep{warner-etal-2025-smarter} as the discriminator. For each task intent, we sample one simulated conversation (using the user and assistant models) and one real conversation (from the ground truth dataset).\footnote{We use only the user turns for discrimination, as we found including assistant turns made simulated conversations perfectly discriminable due to stylistic differences between assistants.} This ensures that our dataset is always balanced. We re-train our discriminator every $25$ steps of RL training. At each iteration, half of our training dataset consists of conversations based on task intents from the most recent $25$ RL steps, while the other half is sampled uniformly from the remaining training history. We train for five epochs and pick the checkpoint achieving the best validation accuracy. Refer to Appendix \ref{sec:appendix_discriminator_training} for exact hyperparameters and training details.

%% file: sections/30_main_evaluation.tex
\subsection{ Baselines}\label{sec:eval_models_and_dataset}
We compare \modelname~ against two categories of models: assistant LLMs prompted to behave like users, and specialized finetuned user simulators, as well as held out reference users. 

\paragraph{Finetuned Models.} We compare against USP~\citep{wang-etal-2025-know}, UserLM-8B~\citep{naous2026flipping}, \modelnamesft{} (our SFT initialization), SFTUser-14B~\citep{suh2026quantifyingutilityusersimulators} and HumanLM-Opinion~\citep{wu2026humanlm}. UserLM-8B and SFTUser-14B are trained on WildChat-1M~\citep{zhao2024wildchat} via SFT but differ in their backbone initialization. USP is also SFT-trained but on a subset of 94,874 LMSys-Chat conversations~\citep{zheng2024lmsyschatm}. HumanLM is trained via RL to align with latent states such as user goals, opinions or emotional valence. 
We use the HumanLM-Opinion checkpoint as other HumanLM checkpoints have not been released. For USP and HumanLM-Opinion, we include the input intent and termination instructions in system prompts (details in Appendix \ref{sec:appendix_finetuned_baseline_prompting}).

\paragraph{Prompted assistant LLMs.} We use a range of small and large, open- and closed-source models. Specifically, we evaluate: Llama3.1-8B-Instruct~\citep{grattafiori2024llama} (Llama-3.1-8B in the tables), Qwen3.5-9B~\citep{qwen35blog}, GPT-5.6-Luna, GPT-5.6-Sol~\citep{openai2026gpt56}, DeepSeek-V4-Pro~\citep{xu2026deepseek} and Kimi-K2.6~\citep{moonshot2026kimik26}. We follow the prompting strategy of \citet{naous2026flipping}, with prompts and inference details in Appendix \ref{sec:appendix_assistant_baseline_prompting}.

\subsection{Evaluation Metrics}\label{sec:eval_metrics} 
We evaluate all simulators on a common set of 2,000 test task intents. For each task intent, we simulate four independent conversations and compute three evaluation metrics that assess task fidelity and three metrics that assess naturalness. Finally, we define a compound metric that assesses the overall \textit{validity} of each simulated conversation.

\paragraph{Task Fidelity Metrics.} We assess each generated conversation by reusing the Terminated, Quality, and Coverage as defined in Section~\ref{sec:reward-design}. In short, Terminated is the percentage of conversations with an explicit \texttt{<|endconversation|>} present. The Quality metric is a binary check for three undesirable behaviors (e.g., verbatim copying of the task intent in a single user utterance). For the evaluation, we used Qwen3.5-9B as the LLM judge for Quality, and the stronger GPT-5.6-Luna for Coverage, as we found this additional model capacity allowed to extract additional per-turn information about intent coverage. 

\paragraph{Naturalness Metrics.} We report naturalness using three complementary discriminators. We trained the first two discriminators in analogous ways to discriminators trained in Section~\ref{sec:training_details}. Both use ModernBert-Large as the base.

\hspace{3mm}\textbf{1. Naturalness-Specific:} We train a separate naturalness discriminator for each simulator to evaluate. The metric reports the average $P(y=\text{real} \mid \hat{u_1} ... \hat{u_k} )$, where $\{\hat{u_1} ... \hat{u_k}\}$ is a simulated conversation on the test set.\footnote{Note that unlike discriminator training in Section \ref{sec:training_details}, we do not include input intents $c$ during discriminator training to avoid doubly penalizing conversations for coverage.} This is the strictest test for naturalness among our metrics as each discriminator specializes on patterns of individual simulators. For Human Reference interactions, we report average $P(y=\text{real} \mid u_1 ... u_k )$ for real conversations $\{u_1 ... u_k\}$, averaged over all model-specific discriminators.
    
\hspace{3mm} \textbf{2. Naturalness-General:} We train a general discriminator on data from all simulators, which includes our models, baselines from \S\ref{sec:experimental_setup}, and experiments with reward design alternatives in \S\ref{sec:analysis_naturalness_formulation}.  
This discriminator is generally less strict than the previous one, as it must distinguish between simulations from multiple distributions.
    
\hspace{3mm} \textbf{3. Pangram:} As a dataset-agnostic discriminator, we use Pangram-4~\citep{glickenhaus2026pangram}, a state-of-the-art AI detection model trained to distinguish human and AI assistant text. We report Pangram's recorded fraction of human-written text. For each intent, we randomly select one sampled conversation. To take into account's Pangram's limitations around short texts, we remove termination tags and concatenate a conversation's user turns through the first whole turn that exceeds a cumulative $100$ words, or use all user turns if the conversation is shorter. 

\paragraph{Naturalness Pass Rate (\%).} A conversation passes the overall naturalness criterion if at least one of the individual naturalness scores (specific, general, Pangram) is at or above $0.5$. This pass rate represents the proportion of simulated conversations where at least one discriminator assigns majority probability to the conversation being classified as real (i.e., the discriminator is ``fooled'').

\paragraph{Conversation Validity Metric.} We define an aggregate metric that measures whether a conversation passes \textit{both} task fidelity and naturalness expectations. This single, over-arching metric captures the proportion of conversations produced by a simulator that are usable for downstream purposes. Under our definition, a simulated conversation is valid if it meets four conditions: (1) a Quality score of $1$, (2) a Terminated score of $1$,\footnote{Human Reference conversations count as terminated by construction.} (3) a Coverage score of at least $0.8$, and (4) passing the naturalness criterion.  We also report estimated cost in USD per 100 valid conversations (lower is better), with base-model pricing proxies for fine-tuned models.

\section{Results and Analysis}\vspace{-2mm}

\begin{table}[t]
\centering
\small
\setlength{\tabcolsep}{3pt}
\renewcommand{\arraystretch}{1.12}
\vspace{-10pt}
\begin{minipage}{0.85\textwidth}
\input{tables/main}
\end{minipage}\par
\vspace{-6pt}
\caption{Evaluation results of user simulators. We compare prompted assistants (top), finetuned models (middle), and human reference conversations on task fidelity (Qual., Cov., Term.) and naturalness (component means Spec., Gen., Pan., and Naturalness Pass Rate, \% Pass). We report overall conversation validity (Valid \%) and estimated simulator cost in USD per 100 valid conversations.} \vspace{-6pt}
\label{tab:main}
\end{table}

\subsection{Main Results}\label{sec:main_results} \vspace{-2mm}

We present summarized evaluation metrics for task fidelity, naturalness, validity, and estimated simulation costs for \modelname{} and baselines in Table~\ref{tab:main}.

\textbf{Comparing prompted assistants and finetuned user simulators.} A general pattern emerges visually in Table~\ref{tab:main}. Prompted assistants tend to achieve high coverage scores, but struggle to generate natural conversations.  
For example, the best prompted assistant (GPT-5.6-Sol) achieves a naturalness pass rate of only 10.2\%.
In contrast, most finetuned user simulators generally perform worse on coverage, but achieve much higher naturalness pass rates. \modelname~achieves optimal performance on both: it achieves the highest performance among simulators on coverage (88.0\%) and naturalness pass rate (71.8\%).

\textbf{\modelname~generates the highest proportion of valid simulations, outperforming the next-best model by $\bm{18.4\%}$.} \modelname~achieves an overall validity score of 46.2\%, which is 1.66 times as much as the next-best model (our finetuned \modelnamesft{}, at a validity score of 27.8\%) and 3.28 times as much as the best previously released user simulator (SFTUser-14B, at 14.1\%). This means that, on average, \modelname~ typically requires sampling two conversations to produce a valid one, when previous models required sampling at least 6, and often more than 15. 
This, coupled with the small parameter count of UserLM-2.0, leads it to be the most cost-effective user simulator, costing \$0.11 to generate 100 valid conversations, compared to \$205.20 with GPT-5.6-Sol.

\textbf{A large gap remains between simulated and real human–AI conversations.} The real, held-out human-AI conversations have perfect naturalness and near-perfect coverage, a 28.2\% and 8.0\% lead over the best user simulator (\modelname{}) on the two metrics respectively. This puts our results in perspective: \modelname{} represents a significant step forward in user simulation, but there remains a substantial gap for future work to close. In particular, our RL-based methodology leads to a large improvement in the task fidelity of the initial SFT checkpoint without sacrificing its naturalness. However, improving beyond this initial naturalness to close the gap to real conversations remains an open challenge. Beyond the model artifact we release, we hope our evaluation methodology emphasizing validity of individual interactions can support future research in user simulation.

\subsection{Stress Testing Under Increasing Task Complexity}\label{sec:stress_test}

\textbf{Setup.} The key use-case for user simulators is supporting downstream evaluation and training. In particular, a user simulator must be capable of simulating realistic and valid interactions as task complexity varies. For evaluation purposes, we decomposed each test task intent into a set of \textit{subgoals}: units of information that are part of the task intent. We argue that the number of subgoals of a task intent roughly correlates with task complexity, as more subgoals typically require longer conversation to convey, leading to more opportunity for user simulators to omit or contradict task information. As a first analysis, we break down the performance of user simulators as the number of subgoals increases. Table~\ref{tab:stress_test_validity} reports coverage, naturalness pass rate, and overall validity against different levels of task intent complexity. We include similar tables for quality, termination, and individual naturalness scores in Appendix~\ref{sec:appendix_stress_test}.

\textbf{\modelname{} maintains the highest validity across all task-complexity levels.} The proportion of valid simulations generally declines as task complexity increases, but \modelname{} leads all evaluated simulators at every level. At five sub-goals, it achieves 36.6\% validity, compared with 11.3\% for \modelnamesft{}. At six and seven sub-goals, \modelname{} retains validity rates of 23.6\% and 12.6\%, compared to below 1\% for every prior simulator.

\begin{table}[t]
\centering
\small
\setlength{\tabcolsep}{2.5pt}
\setlength{\tablegroupsep}{12pt}
\renewcommand{\arraystretch}{1.12}
\vspace{-14pt}
\input{tables/stress_test_validity}
\vspace{-6pt}
\caption{Stress testing by number of task subgoals (1--7). We report mean Coverage, Naturalness Pass Rate (Naturalness Pass \%), and overall Validity. Validity additionally requires quality and termination criteria. Scores are rounded to whole percentages.} \vspace{-6pt}
\label{tab:stress_test_validity}
\end{table}

\textbf{Strong coverage alone does not ensure valid simulations.} \modelname{} achieves near-perfect coverage for single-sub-goal tasks (99.4\%) and the highest coverage among simulators for tasks with more than three sub-goals (Table~\ref{tab:stress_test_validity}). At seven sub-goals, it retains 79.4\% coverage, compared with 78.0\% for GPT-5.6-Luna, the strongest prompted assistant in this bucket, and 58.9\% for \modelnamesft{}. Yet GPT-5.6-Luna produces no valid simulations in this bucket, versus 12.6\% for \modelname{}, illustrating why coverage and naturalness provide complementary views of performance.

\subsection{Behavioral Analysis of Simulators}\label{sec:behavioral_analysis}

\begin{wraptable}[29]{r}{0.5\textwidth}
\vspace{-4ex}
\centering
\small
\setlength{\tabcolsep}{2.5pt}
\renewcommand{\arraystretch}{1.12}
\input{tables/conversational_behavior}
\vspace{-2ex}
\caption{Repetition and verbosity of user simulators. We report the rate of goal revisitation (Revisit), repetition-only turns (Rep. Only), goalless turns that do not advance the task intent (Emp.), and median user words per covered goal (W/goal). Rates are percentages; darker shading indicates closer agreement with humans for each metric.}
\label{tab:conversational_behavior}
\end{wraptable}

For analysis purposes, we recorded which subgoal is revealed at which turn of simulated test conversations, offering a granular turn-by-turn signal to compare simulated conversation dynamics with real conversations. This allows us to perform three targeted analyses of simulator behavior in terms of repetition, verbosity, and turn dynamics.

Table~\ref{tab:conversational_behavior} reports model performance on two repetition and two verbosity metrics (see Appendix~\ref{sec:appendix_behavior_metrics} for exact definitions).

\textbf{Repetition.} First off, human reference conversations tend to include a non-trivial amount of repetition, with 38.8\% of subgoals repeated within a conversation, on average, and 33.3\% of user turn utterances solely repeating information already provided. We find that \modelname{} nearly matches the rates of information revisiting and repetition of human reference conversation (39.0\% and 33.3\%), whereas prompted assistants tend to repeat information (41.8--64.9\%) and have non-progressing turns (35.8--57.4\%) much more frequently. 

\textbf{Verbosity.} We label a turn as \textit{Empty} if it does not cover any subgoal. Such empty turns are relatively rare in real user conversations (6.9\%) but are much more common in most finetuned simulators 18.6--50.5\%, indicative of the tendency of finetuned simulators to go off script. \modelname{} achieves low rates of empty turns (4.3\%), comparable to human reference conversations and strong prompted assistants (3.6\% for GPT-5.6-Sol). Looking at verbosity, we calculate the median number of words each simulator uses on average to reveal a subgoal (W/goal). Human reference conversations are remarkably concise (34 words/subgoal), 40\% less than any simulator we tested. \modelnamesft{} and \modelname{} use a median of 54 and 59 words, respectively, significantly lower than all other simulators (77--315 words per subgoal). Thus, though \modelname{} represents significant progress in terms of lowered verbosity, there remains a large gap with real users.

\begin{figure}[t]
\centering
\includegraphics[width=\textwidth]{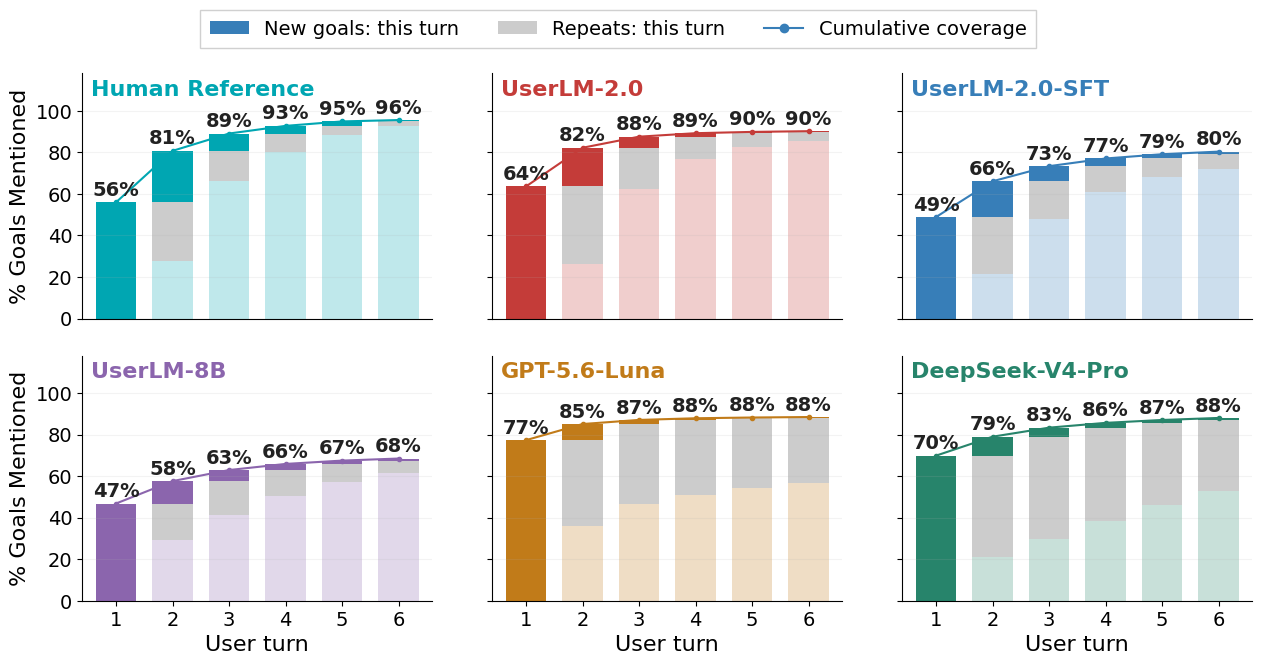}
\caption{We depict goal disclosure over time for the Human Reference, \modelname,and prompted and finetuned baselines. Each bar represents the average cumulative coverage at a given conversation turn across sampled conversations. For each turn, the bolded component represents the proportion of newly covered sub-goals, while the gray component represents the proportion of revisited sub-goals. \modelname~achieves the best alignment with the Human Reference.}
\label{fig:pacing_analysis}
\end{figure}

\textbf{Goal-disclosure.} As a third analysis, we plot coverage progression over time. To enable direct comparisons against human reference conversations, we restrict our focus to conversations with at most six turns. For each turn, we compute the average cumulative coverage and proportion of newly introduced and repeated subgoals at that turn, which we plot in Figure~\ref{fig:pacing_analysis}. Several patterns become apparent: first, when compared to human reference conversations, prompted assistants (GPT-5.6-Luna and DeepSeek-V4-Pro) tend to \textit{front-load} task intent (i.e., reveal a much larger fraction of the entire task intent in the first turn) much more than real conversations (56\% coverage at turn one in real conversations, vs. 70-77\% for prompted assistants). Second, prompted assistants tend to repeat information excessively, whereas repetition is much sparser in real conversations. Finetuned simulators (\modelnamesft{} and UserLM-8B) do not exhibit the same behavior of overly frontloading and repeating subgoals, but they do so at the cost of not achieving high coverage by the end of the conversation (80\% and 68\% coverage by end of conversation, respectively), affecting task fidelity and simulation validity. Finally, conversations simulated with \modelname{} resemble real conversations the most in terms of conversation dynamics: less front-loading and repetition of information, while achieving the highest coverage by the end of the conversation (90\%).

%% file: tables/main.tex
\resizebox{\linewidth}{!}{%
\begin{tabular}{@{}l/*{3}{r}/*{4}{r}/*{2}{r}@{}}
 & \multicolumn{3}{c!{\makebox[\tablegroupsep]{\large \&}}}{\large Task Fidelity} & \multicolumn{4}{c!{\makebox[\tablegroupsep]{\large =}}}{\large Naturalness} & \multicolumn{2}{c}{\large Validity} \\
\cmidrule(r){2-4} \cmidrule(r){5-8} \cmidrule{9-10}
\textbf{Model} & Qual. & Cov. & Term. & Spec. & Gen. & Pan. & \% Pass & Valid \% & Cost \\
\cmidrule(r){1-1}\cmidrule(r){2-4}\cmidrule(r){5-8}\cmidrule{9-10}
DeepSeek-V4-Pro & \cellcolor[HTML]{CDE0EF}98.9 & \cellcolor[HTML]{DFEBF5}85.1 & \cellcolor[HTML]{FFFFFF}21.8 & \cellcolor[HTML]{FEFEFE}0.6 & \cellcolor[HTML]{FEFEFE}2.3 & \cellcolor[HTML]{FFFFFF}0.4 & \cellcolor[HTML]{FFFFFF}2.1 & \cellcolor[HTML]{FFFFFF}0.3 & \$1331.35 \\
Llama-3.1-8B & \cellcolor[HTML]{F4F8FB}59.2 & \cellcolor[HTML]{E7F0F7}76.9 & \cellcolor[HTML]{F9FBFD}48.5 & \cellcolor[HTML]{FFFFFF}0.2 & \cellcolor[HTML]{FEFFFE}1.9 & \cellcolor[HTML]{FDFEFD}3.1 & \cellcolor[HTML]{FEFEFE}2.6 & \cellcolor[HTML]{FFFFFF}0.4 & \$11.98 \\
Qwen3.5-9B & \cellcolor[HTML]{E4EEF6}80.4 & \cellcolor[HTML]{E5EFF7}79.2 & \cellcolor[HTML]{D8E7F2}91.1 & \cellcolor[HTML]{FEFEFE}0.5 & \cellcolor[HTML]{FFFFFF}1.3 & \cellcolor[HTML]{FEFEFE}2.1 & \cellcolor[HTML]{FEFEFE}2.8 & \cellcolor[HTML]{FEFEFF}1.6 & \$14.79 \\
GPT-5.6-Luna & \cellcolor[HTML]{CDE0EF}98.6 & \cellcolor[HTML]{DEEBF4}85.8 & \cellcolor[HTML]{FAFCFD}45.5 & \cellcolor[HTML]{FEFEFE}0.4 & \cellcolor[HTML]{FEFEFE}2.2 & \cellcolor[HTML]{FDFEFD}3.6 & \cellcolor[HTML]{FDFEFD}4.9 & \cellcolor[HTML]{FDFCFE}3.4 & \$20.89 \\
Kimi-K2.6 & \cellcolor[HTML]{CBDFEE}99.6 & \cellcolor[HTML]{DEEBF4}86.1 & \cellcolor[HTML]{F2F7FB}63.5 & \cellcolor[HTML]{FEFEFE}0.5 & \cellcolor[HTML]{FFFFFF}1.5 & \cellcolor[HTML]{FCFDFD}4.2 & \cellcolor[HTML]{FDFEFD}4.9 & \cellcolor[HTML]{FCFCFE}4.0 & \$63.48 \\
GPT-5.6-Sol & \cellcolor[HTML]{CCE0EE}99.3 & \cellcolor[HTML]{E5EFF6}79.7 & \cellcolor[HTML]{F9FBFD}49.8 & \cellcolor[HTML]{FEFFFE}0.3 & \cellcolor[HTML]{FCFDFD}3.8 & \cellcolor[HTML]{FAFCFA}8.3 & \cellcolor[HTML]{FAFBFA}10.2 & \cellcolor[HTML]{FAF9FD}6.7 & \$205.20 \\
\cmidrule(r){1-1}\cmidrule(r){2-4}\cmidrule(r){5-8}\cmidrule{9-10}
HumanLM-Opinion & \cellcolor[HTML]{F4F8FB}60.2 & \cellcolor[HTML]{F5F9FC}58.2 & \cellcolor[HTML]{FFFFFF}19.6 & \cellcolor[HTML]{FFFFFF}0.0 & \cellcolor[HTML]{FEFFFE}2.0 & \cellcolor[HTML]{FFFFFF}0.1 & \cellcolor[HTML]{FFFFFF}1.4 & \cellcolor[HTML]{FFFFFF}0.2 & \$125.26 \\
USP & \cellcolor[HTML]{FAFCFD}46.1 & \cellcolor[HTML]{F2F7FB}63.2 & \cellcolor[HTML]{FBFDFE}42.1 & \cellcolor[HTML]{F7FAF7}4.2 & \cellcolor[HTML]{F5F8F5}11.6 & \cellcolor[HTML]{E6EFE7}39.3 & \cellcolor[HTML]{E5EEE6}43.7 & \cellcolor[HTML]{FCFBFD}5.0 & \$0.37 \\
UserLM-8B & \cellcolor[HTML]{F9FCFD}47.6 & \cellcolor[HTML]{F1F7FB}64.2 & \cellcolor[HTML]{FCFDFE}37.5 & \cellcolor[HTML]{E6EFE7}12.7 & \cellcolor[HTML]{E8F0E9}23.9 & \cellcolor[HTML]{E6EFE7}40.2 & \cellcolor[HTML]{E0EBE2}50.2 & \cellcolor[HTML]{F6F3FA}13.0 & \$0.16 \\
SFTUser-14B & \cellcolor[HTML]{F6FAFC}55.5 & \cellcolor[HTML]{F3F8FB}60.7 & \cellcolor[HTML]{F0F6FA}66.4 & \cellcolor[HTML]{EAF1EB}10.7 & \cellcolor[HTML]{EDF4EE}19.0 & \cellcolor[HTML]{EBF2EC}32.0 & \cellcolor[HTML]{E6EFE7}40.9 & \cellcolor[HTML]{F5F2FA}14.1 & \$2.14 \\
\modelnamesft{} & \cellcolor[HTML]{E8F1F8}75.9 & \cellcolor[HTML]{E8F1F8}76.0 & \cellcolor[HTML]{F2F7FB}62.5 & \cellcolor[HTML]{CDDFCF}25.3 & \cellcolor[HTML]{CDDFCF}50.7 & \cellcolor[HTML]{DFEAE0}50.7 & \cellcolor[HTML]{D5E4D7}68.2 & \cellcolor[HTML]{EBE6F5}27.8 & \$0.19 \\
\modelname{} & \cellcolor[HTML]{D9E8F3}89.8 & \cellcolor[HTML]{DCE9F3}88.0 & \cellcolor[HTML]{E1EDF5}82.8 & \cellcolor[HTML]{EDF3ED}9.4 & \cellcolor[HTML]{DCE8DD}36.1 & \cellcolor[HTML]{D9E7DB}59.2 & \cellcolor[HTML]{D3E3D5}71.8 & \cellcolor[HTML]{DED5EE}46.2 & \$0.11 \\
\cmidrule(r){1-1}\cmidrule(r){2-4}\cmidrule(r){5-8}\cmidrule{9-10}
Human Reference & 87.9 & 96.0 & 100.0 & 96.4 & 84.3 & 85.1 & 100.0 & 80.2 & -- \\
\cmidrule[\heavyrulewidth](r){1-1}\cmidrule[\heavyrulewidth](r){2-4}\cmidrule[\heavyrulewidth](r){5-8}\cmidrule[\heavyrulewidth]{9-10}
\end{tabular}%
}

%% file: tables/stress_test_validity.tex
\resizebox{\linewidth}{!}{%
\begin{tabular}{@{}l/*{7}{r}/*{7}{r}/*{7}{r}@{}}
 & \multicolumn{7}{c}{\large Coverage} & \multicolumn{7}{c}{\large Naturalness Pass (\%)} & \multicolumn{7}{c}{\large Overall Valid \%} \\
\cmidrule(r){2-8} \cmidrule(r){9-15} \cmidrule{16-22}
\textbf{Model} & 1 & 2 & 3 & 4 & 5 & 6 & 7 & 1 & 2 & 3 & 4 & 5 & 6 & 7 & 1 & 2 & 3 & 4 & 5 & 6 & 7 \\
\cmidrule(r){1-1}\cmidrule(r){2-8}\cmidrule(r){9-15}\cmidrule{16-22}
DeepSeek-V4-Pro & \cellcolor[HTML]{CFE2EF}95 & \cellcolor[HTML]{D3E4F1}91 & \cellcolor[HTML]{D5E5F1}88 & \cellcolor[HTML]{D7E7F2}85 & \cellcolor[HTML]{DBE9F3}81 & \cellcolor[HTML]{DEEBF4}78 & \cellcolor[HTML]{E1ECF5}75 & \cellcolor[HTML]{FCFDFD}5 & \cellcolor[HTML]{FDFEFD}4 & \cellcolor[HTML]{FEFEFE}2 & \cellcolor[HTML]{FEFFFE}1 & \cellcolor[HTML]{FFFFFF}1 & \cellcolor[HTML]{FEFFFE}1 & \cellcolor[HTML]{FFFFFF}1 & \cellcolor[HTML]{FEFEFF}2 & \cellcolor[HTML]{FFFFFF}0 & \cellcolor[HTML]{FFFFFF}0 & \cellcolor[HTML]{FFFFFF}0 & \cellcolor[HTML]{FFFFFF}0 & \cellcolor[HTML]{FFFFFF}0 & \cellcolor[HTML]{FFFFFF}0 \\
Llama-3.1-8B & \cellcolor[HTML]{CEE1EF}96 & \cellcolor[HTML]{D6E6F2}86 & \cellcolor[HTML]{DAE8F3}82 & \cellcolor[HTML]{E0ECF5}76 & \cellcolor[HTML]{E5EFF7}70 & \cellcolor[HTML]{E9F1F8}65 & \cellcolor[HTML]{EEF5FA}59 & \cellcolor[HTML]{FBFDFB}7 & \cellcolor[HTML]{FDFEFD}4 & \cellcolor[HTML]{FDFEFD}4 & \cellcolor[HTML]{FEFFFE}1 & \cellcolor[HTML]{FFFFFF}0 & \cellcolor[HTML]{FEFEFE}2 & \cellcolor[HTML]{FFFFFF}1 & \cellcolor[HTML]{FEFEFF}1 & \cellcolor[HTML]{FFFFFF}1 & \cellcolor[HTML]{FFFFFF}1 & \cellcolor[HTML]{FFFFFF}1 & \cellcolor[HTML]{FFFFFF}0 & \cellcolor[HTML]{FFFFFF}0 & \cellcolor[HTML]{FFFFFF}0 \\
Qwen3.5-9B & \cellcolor[HTML]{D3E4F1}90 & \cellcolor[HTML]{D5E5F1}88 & \cellcolor[HTML]{D9E8F3}83 & \cellcolor[HTML]{DEEBF4}78 & \cellcolor[HTML]{E1EDF5}74 & \cellcolor[HTML]{E4EEF6}71 & \cellcolor[HTML]{E9F1F8}66 & \cellcolor[HTML]{F9FBF9}11 & \cellcolor[HTML]{FDFEFD}4 & \cellcolor[HTML]{FEFEFE}2 & \cellcolor[HTML]{FEFEFE}2 & \cellcolor[HTML]{FFFFFF}1 & \cellcolor[HTML]{FFFFFF}0 & \cellcolor[HTML]{FFFFFF}0 & \cellcolor[HTML]{FCFBFD}9 & \cellcolor[HTML]{FEFEFF}2 & \cellcolor[HTML]{FFFFFF}1 & \cellcolor[HTML]{FFFFFF}1 & \cellcolor[HTML]{FFFFFF}0 & \cellcolor[HTML]{FFFFFF}0 & \cellcolor[HTML]{FFFFFF}0 \\
GPT-5.6-Luna & \cellcolor[HTML]{CDE0EF}97 & \cellcolor[HTML]{D3E4F1}90 & \cellcolor[HTML]{D5E5F1}88 & \cellcolor[HTML]{D7E6F2}86 & \cellcolor[HTML]{DAE8F3}83 & \cellcolor[HTML]{DDEAF4}79 & \cellcolor[HTML]{DEEBF4}78 & \cellcolor[HTML]{F4F8F4}22 & \cellcolor[HTML]{FAFCFB}9 & \cellcolor[HTML]{FEFEFE}3 & \cellcolor[HTML]{FEFFFE}1 & \cellcolor[HTML]{FFFFFF}1 & \cellcolor[HTML]{FEFFFE}1 & \cellcolor[HTML]{FFFFFF}1 & \cellcolor[HTML]{F7F5FB}21 & \cellcolor[HTML]{FDFDFE}5 & \cellcolor[HTML]{FEFEFF}2 & \cellcolor[HTML]{FFFFFF}1 & \cellcolor[HTML]{FFFFFF}0 & \cellcolor[HTML]{FFFFFF}0 & \cellcolor[HTML]{FFFFFF}0 \\
Kimi-K2.6 & \cellcolor[HTML]{CEE1EF}96 & \cellcolor[HTML]{D1E3F0}92 & \cellcolor[HTML]{D4E4F1}89 & \cellcolor[HTML]{D7E6F2}86 & \cellcolor[HTML]{DAE8F3}82 & \cellcolor[HTML]{DDEAF4}79 & \cellcolor[HTML]{E0ECF5}76 & \cellcolor[HTML]{F4F8F4}22 & \cellcolor[HTML]{FAFCFA}10 & \cellcolor[HTML]{FDFEFD}3 & \cellcolor[HTML]{FEFFFE}1 & \cellcolor[HTML]{FFFFFF}0 & \cellcolor[HTML]{FFFFFF}1 & \cellcolor[HTML]{FFFFFF}0 & \cellcolor[HTML]{F7F5FB}22 & \cellcolor[HTML]{FCFBFD}9 & \cellcolor[HTML]{FFFFFF}1 & \cellcolor[HTML]{FFFFFF}1 & \cellcolor[HTML]{FFFFFF}0 & \cellcolor[HTML]{FFFFFF}0 & \cellcolor[HTML]{FFFFFF}0 \\
GPT-5.6-Sol & \cellcolor[HTML]{CEE1EF}96 & \cellcolor[HTML]{D5E5F1}88 & \cellcolor[HTML]{DAE8F3}82 & \cellcolor[HTML]{DDEAF4}79 & \cellcolor[HTML]{E1ECF5}75 & \cellcolor[HTML]{E5EFF7}70 & \cellcolor[HTML]{E7F0F7}68 & \cellcolor[HTML]{ECF3ED}36 & \cellcolor[HTML]{F5F9F6}18 & \cellcolor[HTML]{FAFBFA}11 & \cellcolor[HTML]{FDFEFD}4 & \cellcolor[HTML]{FEFFFE}1 & \cellcolor[HTML]{FDFEFD}3 & \cellcolor[HTML]{FEFEFE}3 & \cellcolor[HTML]{F2EFF9}35 & \cellcolor[HTML]{FBFAFD}11 & \cellcolor[HTML]{FDFDFE}5 & \cellcolor[HTML]{FEFEFF}1 & \cellcolor[HTML]{FFFFFF}1 & \cellcolor[HTML]{FFFFFF}0 & \cellcolor[HTML]{FFFFFF}0 \\
\cmidrule(r){1-1}\cmidrule(r){2-8}\cmidrule(r){9-15}\cmidrule{16-22}
HumanLM-Opinion & \cellcolor[HTML]{F2F7FB}56 & \cellcolor[HTML]{EEF5FA}59 & \cellcolor[HTML]{EDF4F9}61 & \cellcolor[HTML]{EEF5FA}59 & \cellcolor[HTML]{EFF5FA}59 & \cellcolor[HTML]{F2F7FB}56 & \cellcolor[HTML]{F3F7FB}54 & \cellcolor[HTML]{FDFEFD}4 & \cellcolor[HTML]{FEFEFE}3 & \cellcolor[HTML]{FFFFFF}1 & \cellcolor[HTML]{FEFFFE}1 & \cellcolor[HTML]{FFFFFF}1 & \cellcolor[HTML]{FFFFFF}0 & \cellcolor[HTML]{FFFFFF}0 & \cellcolor[HTML]{FFFFFF}0 & \cellcolor[HTML]{FFFFFF}0 & \cellcolor[HTML]{FFFFFF}0 & \cellcolor[HTML]{FFFFFF}1 & \cellcolor[HTML]{FFFFFF}0 & \cellcolor[HTML]{FFFFFF}0 & \cellcolor[HTML]{FFFFFF}0 \\
USP & \cellcolor[HTML]{DCEAF4}80 & \cellcolor[HTML]{E6EFF7}69 & \cellcolor[HTML]{EAF2F8}65 & \cellcolor[HTML]{EDF4F9}61 & \cellcolor[HTML]{EEF5FA}59 & \cellcolor[HTML]{F0F6FA}57 & \cellcolor[HTML]{F5F9FC}51 & \cellcolor[HTML]{E1ECE2}58 & \cellcolor[HTML]{E7F0E8}46 & \cellcolor[HTML]{E5EEE6}50 & \cellcolor[HTML]{E9F1EA}42 & \cellcolor[HTML]{EAF2EB}40 & \cellcolor[HTML]{EBF2EC}38 & \cellcolor[HTML]{F0F5F0}29 & \cellcolor[HTML]{FAF8FC}14 & \cellcolor[HTML]{FCFBFD}10 & \cellcolor[HTML]{FDFDFE}5 & \cellcolor[HTML]{FEFEFE}3 & \cellcolor[HTML]{FEFEFF}3 & \cellcolor[HTML]{FFFFFF}0 & \cellcolor[HTML]{FFFFFF}1 \\
UserLM-8B & \cellcolor[HTML]{D5E5F1}88 & \cellcolor[HTML]{E1EDF5}74 & \cellcolor[HTML]{E9F1F8}65 & \cellcolor[HTML]{ECF3F9}62 & \cellcolor[HTML]{F0F6FA}57 & \cellcolor[HTML]{F3F8FB}54 & \cellcolor[HTML]{F9FBFD}47 & \cellcolor[HTML]{D8E6D9}76 & \cellcolor[HTML]{DCE9DE}66 & \cellcolor[HTML]{E0EBE2}59 & \cellcolor[HTML]{E5EFE6}49 & \cellcolor[HTML]{EBF2EC}39 & \cellcolor[HTML]{EFF5F0}30 & \cellcolor[HTML]{F0F5F0}29 & \cellcolor[HTML]{EDE8F6}50 & \cellcolor[HTML]{F5F2FA}28 & \cellcolor[HTML]{FCFBFD}10 & \cellcolor[HTML]{FEFEFE}3 & \cellcolor[HTML]{FDFDFE}5 & \cellcolor[HTML]{FFFFFF}1 & \cellcolor[HTML]{FFFFFF}0 \\
SFTUser-14B & \cellcolor[HTML]{D5E5F1}88 & \cellcolor[HTML]{E2EDF6}73 & \cellcolor[HTML]{EAF2F8}64 & \cellcolor[HTML]{F0F6FA}57 & \cellcolor[HTML]{F4F8FB}53 & \cellcolor[HTML]{F9FBFD}47 & \cellcolor[HTML]{FFFFFF}40 & \cellcolor[HTML]{D8E6DA}74 & \cellcolor[HTML]{DFEAE0}62 & \cellcolor[HTML]{EAF1EB}41 & \cellcolor[HTML]{F0F5F1}29 & \cellcolor[HTML]{F0F5F0}30 & \cellcolor[HTML]{F1F6F1}27 & \cellcolor[HTML]{F1F6F1}28 & \cellcolor[HTML]{EAE5F4}56 & \cellcolor[HTML]{F3F0F9}33 & \cellcolor[HTML]{FCFBFD}9 & \cellcolor[HTML]{FEFEFF}3 & \cellcolor[HTML]{FEFDFE}4 & \cellcolor[HTML]{FFFFFF}1 & \cellcolor[HTML]{FFFFFF}1 \\
\modelnamesft{} & \cellcolor[HTML]{CDE0EF}97 & \cellcolor[HTML]{D6E6F2}87 & \cellcolor[HTML]{DEEBF4}78 & \cellcolor[HTML]{E1EDF5}74 & \cellcolor[HTML]{E5EFF7}70 & \cellcolor[HTML]{EAF2F8}64 & \cellcolor[HTML]{EFF5FA}59 & \cellcolor[HTML]{CDDFCF}96 & \cellcolor[HTML]{D0E1D2}90 & \cellcolor[HTML]{D9E7DA}74 & \cellcolor[HTML]{DDE9DE}65 & \cellcolor[HTML]{E1ECE2}58 & \cellcolor[HTML]{E6EFE7}49 & \cellcolor[HTML]{EBF2EC}38 & \cellcolor[HTML]{E0D8EF}85 & \cellcolor[HTML]{EAE4F4}58 & \cellcolor[HTML]{F6F4FB}24 & \cellcolor[HTML]{FAF8FC}15 & \cellcolor[HTML]{FBFAFD}11 & \cellcolor[HTML]{FDFCFE}6 & \cellcolor[HTML]{FEFEFE}3 \\
\modelname{} & \cellcolor[HTML]{CBDFEE}99 & \cellcolor[HTML]{D0E2F0}94 & \cellcolor[HTML]{D5E5F1}88 & \cellcolor[HTML]{D5E5F1}88 & \cellcolor[HTML]{D8E7F2}85 & \cellcolor[HTML]{DAE8F3}82 & \cellcolor[HTML]{DDEAF4}79 & \cellcolor[HTML]{CEDFD0}95 & \cellcolor[HTML]{D2E2D4}86 & \cellcolor[HTML]{D6E5D8}79 & \cellcolor[HTML]{DCE8DD}68 & \cellcolor[HTML]{DEEADF}64 & \cellcolor[HTML]{E2EDE3}55 & \cellcolor[HTML]{E3EDE4}53 & \cellcolor[HTML]{DED5EE}91 & \cellcolor[HTML]{E4DDF1}74 & \cellcolor[HTML]{EDE8F6}50 & \cellcolor[HTML]{F3F0F9}32 & \cellcolor[HTML]{F2EEF8}37 & \cellcolor[HTML]{F6F4FB}24 & \cellcolor[HTML]{FAF9FD}13 \\
\cmidrule(r){1-1}\cmidrule(r){2-8}\cmidrule(r){9-15}\cmidrule{16-22}
Human Reference & 100 & 97 & 93 & 93 & 96 & 97 & 99 & 100 & 100 & 100 & 100 & 100 & 100 & 100 & 88 & 90 & 76 & 68 & 82 & 79 & 82 \\
\cmidrule[\heavyrulewidth](r){1-1}\cmidrule[\heavyrulewidth](r){2-8}\cmidrule[\heavyrulewidth](r){9-15}\cmidrule[\heavyrulewidth]{16-22}
\end{tabular}%
}

%% file: tables/conversational_behavior.tex
\resizebox{\linewidth}{!}{%
\begin{tabular}{@{}l*{4}{r}@{}}
 & \multicolumn{2}{c}{Repetition} & \multicolumn{2}{c}{Verbosity} \\
\cmidrule(lr){2-3} \cmidrule(lr){4-5}
\textbf{Model} & \makecell{Re-\\visit} & \makecell{Rep.\\Only} & Emp. & W/goal \\
\cmidrule(r){1-1}\cmidrule(r){2-3}\cmidrule{4-5}
DeepSeek-V4-Pro & \cellcolor[HTML]{FFFFFF}64.9 & \cellcolor[HTML]{FFFFFF}57.4 & \cellcolor[HTML]{D1E2D3}12.6 & \cellcolor[HTML]{DBE8DD}127 \\
Llama-3.1-8B & \cellcolor[HTML]{FAF0E5}53.1 & \cellcolor[HTML]{F8EBDD}43.2 & \cellcolor[HTML]{DFEBE1}24.2 & \cellcolor[HTML]{E7F0E8}190 \\
Qwen3.5-9B & \cellcolor[HTML]{F4E1CC}41.8 & \cellcolor[HTML]{F4E0CB}35.8 & \cellcolor[HTML]{D4E3D5}14.5 & \cellcolor[HTML]{D6E5D7}99 \\
GPT-5.6-Luna & \cellcolor[HTML]{F9EDE0}51.0 & \cellcolor[HTML]{FAF1E8}47.8 & \cellcolor[HTML]{CDDFCF}8.9 & \cellcolor[HTML]{D5E4D7}96 \\
Kimi-K2.6 & \cellcolor[HTML]{F9EFE4}52.4 & \cellcolor[HTML]{FAF0E7}47.2 & \cellcolor[HTML]{CDDFCF}4.6 & \cellcolor[HTML]{D2E2D3}77 \\
GPT-5.6-Sol & \cellcolor[HTML]{F9EFE4}52.4 & \cellcolor[HTML]{F9EDE1}45.1 & \cellcolor[HTML]{CEE0D0}3.6 & \cellcolor[HTML]{D3E3D4}83 \\
\cmidrule(r){1-1}\cmidrule(r){2-3}\cmidrule{4-5}
HumanLM-Opinion & \cellcolor[HTML]{F6E5D4}45.3 & \cellcolor[HTML]{F5E3D1}28.6 & \cellcolor[HTML]{FFFFFF}50.5 & \cellcolor[HTML]{FFFFFF}315 \\
USP & \cellcolor[HTML]{F5E4D1}33.6 & \cellcolor[HTML]{F9EFE4}20.4 & \cellcolor[HTML]{FAFCFA}46.6 & \cellcolor[HTML]{D3E3D5}86 \\
UserLM-8B & \cellcolor[HTML]{F5E2CE}35.1 & \cellcolor[HTML]{F8EADB}24.0 & \cellcolor[HTML]{E8F0E9}31.2 & \cellcolor[HTML]{EAF1EB}204 \\
SFTUser-14B & \cellcolor[HTML]{FDFAF6}16.7 & \cellcolor[HTML]{FEFDFC}10.3 & \cellcolor[HTML]{E8F1E9}31.8 & \cellcolor[HTML]{DAE7DB}120 \\
\modelnamesft{} & \cellcolor[HTML]{F4DEC8}40.2 & \cellcolor[HTML]{F4DFC9}31.8 & \cellcolor[HTML]{D9E6DA}18.6 & \cellcolor[HTML]{CDDFCF}54 \\
\modelname{} & \cellcolor[HTML]{F3DDC6}39.0 & \cellcolor[HTML]{F3DDC6}33.0 & \cellcolor[HTML]{CEDFD0}4.3 & \cellcolor[HTML]{CEE0D0}59 \\
\cmidrule(r){1-1}\cmidrule(r){2-3}\cmidrule{4-5}
Human Reference & 38.8 & 33.3 & 6.9 & 34 \\
\bottomrule
\end{tabular}%
}

%% file: sections/40_analysis.tex
\subsection{Reward Design Alternatives}\label{sec:analysis_naturalness_formulation}
\begin{wraptable}{R}{0.5\textwidth}
\centering
\small
\setlength{\tabcolsep}{2pt}
\setlength{\tablegroupsep}{4pt}
\renewcommand{\arraystretch}{1.15}
\vspace{-8mm}
\input{tables/ablation_full}
\vspace{-4mm}
\caption{Evaluation results of alternative training strategies. The top three models are trained with alternative reward designs for naturalness. The middle models are either SFT only, or ablate a reward component (Coverage or Naturalness). We report task-fidelity, naturalness and validity metrics.}
\label{tab:ablation}
\end{wraptable}
\textbf{Choice of naturalness discriminators.} First, we explore alternative options for the design of the naturalness reward. We compare the performance of \modelname{}, trained with a dynamic discriminator, against the following choices: (i) Fixed Discriminator, trained only once at the beginning of RL, (ii) LLM Judge - Likert, which prompts an LLM-judge to score naturalness on a Likert scale of 1 to 5, (iii) LLM Judge - Winrate, which prompts the judge to select between a model-generated conversation and its corresponding ground-truth variant and uses the resulting winrate as the reward. Refer to Appendix \ref{sec:appendix_llm_judge} for details.

Table~\ref{tab:ablation} reports task-fidelity, naturalness, and validity metrics for all discriminator options. Our results show that dynamic discriminators are necessary for achieving high naturalness. \modelname, trained with the dynamic discriminator, outperforms all naturalness alternatives by a large margin. We find that the Fixed Disc. alternative reports the lowest naturalness pass rate of any discriminator. Manual inspection of conversations revealed that this user simulator converges to a narrow range of word- and turn-counts that consistently fools the frozen discriminator but does not reflect true naturalness. Lastly, neither user simulator trained with LLM-judge based discriminators come close to \modelname's naturalness. This is unsurprising as naturalness is a distributional property of a corpus which is difficult for LLM-judges to infer based on a few exemplars or instructions.

\paragraph{Reward components.}
Second, we train separate models that remove the naturalness reward (W/o Naturalness) or the coverage reward (W/o Coverage). Since the termination reward described in \S\ref{sec:reward-design} explicitly captures coverage, we also omit it when training the W/o Coverage model. All other reward components remain unchanged.

The middle block of Table~\ref{tab:ablation} reports on these ablations. Both removing naturalness and jointly removing coverage and termination reduce validity. Neither ablated model matches the validity score of \modelname. This reflects the requirement that valid simulations satisfy both task-fidelity and naturalness criteria.

The removal of the naturalness reward (W/o Naturalness) lowers the rate at which the model fools any discriminator (42.0\% vs 71.8\% for the full \modelname). The W/o Coverage model retains a high naturalness pass rate (81.2\%), but achieves poor overall validity due to task-fidelity metrics (19.0\% vs 46.2\% for the full model). This limits its joint validity despite its naturalness. In the component-level task-fidelity evaluations, this model reports a poor Terminated score that checks for the \texttt{<|endconversation|>} tag. A seemingly counterintuitive result is that it still reports high mean coverage. However, recall that our naturalness reward also indirectly captures coverage. The naturalness discriminators during training (see \S\ref{sec:reward-design}) condition on the task intents $c$ and are therefore likely to learn to reward conversations that satisfy them, as this behavior is exhibited by ground-truth conversations.

\section{Discussion \& Implications}

\paragraph{The remaining gap between \modelname{} and real user conversations.} We first note that our existing metrics are imperfect, as human reference conversations do not achieve perfect scores on Quality or Coverage. Nonetheless, the human references still strongly outperform all simulators, including \modelname, on these metrics, despite our training directly optimizing these metrics. According to our framework, more than 80\% of reference conversations are valid in terms of task fidelity and naturalness. This is more than 30\% above \modelname{}, a gap roughly as large as the gain between the previously best simulator and \modelname{} (14.1\% to 46.2\%). We believe that further innovation and scaling in the RL recipes could yield further gains on the metrics within our framework, delivering state-of-the-art user simulators. In other words, we believe the gap in metrics that remains in our framework is significant and invite the community to close the gap to develop improved user simulators.

\paragraph{Poorly chosen naturalness rewards can be worse than none at all.} One surprising finding from our alternative reward design experiments (Section~\ref{sec:analysis_naturalness_formulation}) is that several formulations of naturalness rewards lead to worse naturalness performance than no naturalness reward at all. More specifically, we found that evaluating naturalness either with an LLM-as-a-judge prompt or using a fixed discriminator trained only once can harm simulator performance by leading to simulations that are more easily distinguishable from real conversations (i.e., less natural). This phenomenon is understandable through the lens of reward hacking: the simulator optimizes for a flawed notion of naturalness, making simulated conversations more easily identifiable and less natural. This finding highlights the importance of the dynamically retrained discriminator of our training framework: we demonstrated empirically that this enables the trained simulators to minimize the gap with a reference corpus of real interactions while preventing reward hacking. We hope this experiment encourages the community to limit the use of LLM-as-a-judge for naturalness evaluation, as we show it is not effective.

\paragraph{\modelname~is trained and evaluated only on chatbot interactions.} For \modelname, our training and evaluation intend to simulate conversations from the WildChat-4.8M~\citep{zhao2024wildchat} dataset. WildChat conversations are pure chatbot interactions, with no environment present (such as a file system or access to a computing environment). As agentic interaction with AI-based systems become more prevalent, being able to simulate users not only in pure chat settings but also in agentic settings that involve tools and environments become more important. This can help evaluate not only how users dynamically reveal their intents, but also how they access files, verify task completions from the agent, and how they choose to delegate tasks within a shared environment. Creating such agentic user simulators faces several key challenges, including a lack of diverse data of agentic user-AI interactions. Initial resources such as SWE-chat~\citep{baumann2026swechat} show a promising path forward that can enable the training of user simulators for agentic settings.

\paragraph{Alternative designs for the discriminator.} During training, we made several decisions on the information the naturalness discriminator is conditioned on. Specifically, the discriminator we use sees all user utterances, none of the assistant utterances, and the input task intent for a given simulated conversation. This causes the discriminator to not only assess whether the user utterances are natural in terms of the surface form, but also whether the conversation is a naturalistic enactment of the entire task intent (as real conversations are always full enactments of the task intent). We believe that alternative inputs to the discriminator can define subtly distinct naturalness rewards, which can be combined to produce more robust and well-rounded naturalness rewards for training. For example, one could formulate turn-level or conversation-level discriminators, with or without task intents provided, with or without assistant utterances included. We hope that the community will explore diverse formulations of discriminator conditions, and find effective ways to combine them into robust methods to train natural user simulators.

\paragraph{User simulators for real-world use-cases must take privacy into account.} Naturalness is a question of how well a given user simulator aligns with behavior in a target corpus. \modelname, for instance, is trained by first performing SFT on WildChat-4.8M to align its initial behavior with that of users in the dataset. In realistic settings, simply doing SFT on user data runs the risk of having models regurgitate private information. The development of user simulators should work in tandem with the design of privacy-aware training strategies or methods for generating synthetic data for training that remains naturalistic.

\paragraph{User simulators should be robust to a variety of assistants.} Our training and evaluation both used Qwen3.5-9B~\citep{qwen35blog} as the assistant model that the simulated user interacts with. For a user simulator to be robust, however, it should provide naturalistic responses to both state-of-the-art assistant models, as well as those that perform poorly or inconsistently in a given task. Devising training and evaluation environments that expose user simulators to a broader range of assistant behaviors is a valuable next step.

\paragraph{User simulators should be evaluated holistically.}
Research on user simulators is still nascent, and we currently lack a standardized evaluation methodology. One line of work evaluates simulators by comparing corpus-level statistics of behaviors like turn-counts, turn-lengths, etc. against a corpus of human-AI conversations. These metrics are intuitive and useful for characterizing aggregate behaviors, as shown in our behavioral analysis in Section~\ref{sec:behavioral_analysis}. However, a corpus-level alignment on multiple behaviors independently does not guarantee that individual simulations are usable. For example, consider a simulator that, on aggregate, improves both coverage and naturalness but each generated sample maximizes one at the cost of the other. In this case, no individual sample is usable for downstream use-cases of evaluating or training assistant models. Therefore, in addition to corpus-level metrics, we believe that a simulation-level metric is crucial to establishing simulation readiness. Our validity metric captures this new evaluation principle. In our work, we instantiated it to combine task fidelity and naturalness dimensions, although this exact combination can be altered depending on domain specifications. We hope that future work will evaluate user simulators using both simulation- and corpus-level tests for holistic evaluation.

%% file: tables/ablation_full.tex
\resizebox{\linewidth}{!}{%
\begin{tabular}{@{}l/*{3}{r}/*{4}{r}/*{1}{r}@{}}
 & \multicolumn{3}{c}{Task Fidelity} & \multicolumn{4}{c}{Naturalness} & \multicolumn{1}{c}{Valid} \\
\cmidrule(r){2-4} \cmidrule(r){5-8} \cmidrule{9-9}
\textbf{Model} & Qual. & Cov. & Term. & Spec. & Gen. & Pan. & Pass. & \% \\
\cmidrule(r){1-1}\cmidrule(r){2-4}\cmidrule(r){5-8}\cmidrule{9-9}
\multicolumn{9}{l}{\textit{Naturalness Design}} \\
Fixed Disc. & \cellcolor[HTML]{D1E3F0}95.8 & \cellcolor[HTML]{DBE9F3}88.7 & \cellcolor[HTML]{EAF2F8}73.4 & \cellcolor[HTML]{FEFEFE}0.8 & \cellcolor[HTML]{FDFEFD}3.2 & \cellcolor[HTML]{F9FBF9}10.1 & \cellcolor[HTML]{F9FBF9}11.3 & \cellcolor[HTML]{FAF8FC}7.8 \\
Judge (Likert) & \cellcolor[HTML]{CEE1EF}98.0 & \cellcolor[HTML]{D7E6F2}91.6 & \cellcolor[HTML]{E4EEF6}80.7 & \cellcolor[HTML]{FDFEFD}1.0 & \cellcolor[HTML]{FFFFFF}1.8 & \cellcolor[HTML]{F7FAF7}13.1 & \cellcolor[HTML]{F7FAF7}14.0 & \cellcolor[HTML]{F8F6FB}10.5 \\
Judge (Winrate) & \cellcolor[HTML]{D0E2F0}96.6 & \cellcolor[HTML]{D5E5F1}92.8 & \cellcolor[HTML]{FDFEFE}32.2 & \cellcolor[HTML]{FDFEFD}1.0 & \cellcolor[HTML]{FBFDFC}4.9 & \cellcolor[HTML]{F0F6F1}23.1 & \cellcolor[HTML]{F0F6F1}24.5 & \cellcolor[HTML]{FAF9FD}7.0 \\
\addlinespace[2pt]
\cmidrule(r){1-1}\cmidrule(r){2-4}\cmidrule(r){5-8}\cmidrule{9-9}
\multicolumn{9}{l}{\textit{Reward Ablations}} \\
\modelnamesft{} & \cellcolor[HTML]{E8F1F8}75.9 & \cellcolor[HTML]{E8F1F8}76.0 & \cellcolor[HTML]{F2F7FB}62.5 & \cellcolor[HTML]{CDDFCF}25.3 & \cellcolor[HTML]{CDDFCF}50.7 & \cellcolor[HTML]{DFEAE0}50.7 & \cellcolor[HTML]{D5E4D7}68.2 & \cellcolor[HTML]{EBE6F5}27.8 \\
W/o Naturalness & \cellcolor[HTML]{D4E4F1}94.0 & \cellcolor[HTML]{D8E7F2}90.4 & \cellcolor[HTML]{F6F9FC}56.2 & \cellcolor[HTML]{F2F7F3}6.6 & \cellcolor[HTML]{EBF2EC}21.0 & \cellcolor[HTML]{E9F1EA}34.4 & \cellcolor[HTML]{E6EFE7}42.0 & \cellcolor[HTML]{F0ECF7}20.7 \\
W/o Coverage & \cellcolor[HTML]{DBE9F3}88.2 & \cellcolor[HTML]{D9E8F3}89.8 & \cellcolor[HTML]{FEFEFF}30.3 & \cellcolor[HTML]{FAFCFA}2.7 & \cellcolor[HTML]{F1F6F1}15.3 & \cellcolor[HTML]{CDDFCF}78.8 & \cellcolor[HTML]{CDDFCF}81.2 & \cellcolor[HTML]{F2EEF8}19.0 \\
W/o Termination & \cellcolor[HTML]{E4EFF6}80.0 & \cellcolor[HTML]{D9E7F2}90.4 & \cellcolor[HTML]{FDFEFE}32.6 & \cellcolor[HTML]{FCFDFC}1.8 & \cellcolor[HTML]{F8FAF8}8.5 & \cellcolor[HTML]{D3E3D5}68.7 & \cellcolor[HTML]{D3E3D5}71.0 & \cellcolor[HTML]{F2EFF8}18.1 \\
\addlinespace[2pt]
\cmidrule(r){1-1}\cmidrule(r){2-4}\cmidrule(r){5-8}\cmidrule{9-9}
\modelname{} & \cellcolor[HTML]{D9E8F3}89.8 & \cellcolor[HTML]{DCE9F3}88.0 & \cellcolor[HTML]{E1EDF5}82.8 & \cellcolor[HTML]{EDF3ED}9.4 & \cellcolor[HTML]{DCE8DD}36.1 & \cellcolor[HTML]{D9E7DB}59.2 & \cellcolor[HTML]{D3E3D5}71.8 & \cellcolor[HTML]{DED5EE}46.2 \\
\cmidrule[\heavyrulewidth](r){1-1}\cmidrule[\heavyrulewidth](r){2-4}\cmidrule[\heavyrulewidth](r){5-8}\cmidrule[\heavyrulewidth]{9-9}
\end{tabular}%
}

%% file: sections/50_related_work.tex
\section{Related Work}
\textbf{User Simulation: } Early work focused on user simulation as a means to evaluate or train task-oriented dialogue policies~\citep{658991, shi2019build}, with methods ranging from rule-based systems~\citep{schatzmann-etal-2007-agenda} to neural architectures~\citep{asri16_interspeech, kreyssig-etal-2018-neural}. Currently, user simulation is predominantly done using LLM assistants prompted zero-shot to act as users on settings ranging from chatbot interactions~\citep{li2024iqa,wu2025collabllm} to agentic tasks~\citep{barres2026taubench,sun2026training}. While such simulations benefit from LLMs' ability to enact specified task intents, they also suffer from poor alignment with human behavior~\citep{yoon-etal-2024-evaluating, ivey-etal-2026-real, zhou2026mind}. Prior work has separately explored training models to generate user turns in dialogues. This has most commonly been done through finetuning on human-assistant interactions~\citep{sekulic-etal-2024-reliable, naous2026flipping, suh2026quantifyingutilityusersimulators} on datasets such as MultiWOZ and WildChat~\citep{eric-etal-2020-multiwoz, zhao2024wildchat}. The resulting models are naturalistic, but also brittle and poor at task-fidelity~\citep{naous2026flipping}. As a result, recent work has placed emphasis on alignment and consistency with user intents via RL~\citep{abdulhai2025consistently,wu2026humanlm,mehri2026goal}. We propose training an SFT initialization using RL to maximize both task fidelity and naturalness, operationalized as fooling a dynamically trained discriminator.

\textbf{Evaluation for User Simulators: } There is currently no consensus on how to evaluate user simulators~\citep{pietquin2013survey,deriu2021survey}. One common approach is evaluating similarity to human interactions, which can involve comparisons of user simulator surface-form statistics and stylistic features~\citep{schatzmann2005quantitative,pietquin2013survey,ivey-etal-2026-real,zhou2026mind} or behavioral patterns~\citep{zhou2026mind, mehri2026measuring} against human data. Alternatively, similarity can be assessed by how well user simulators approximate assistant performance with real humans~\citep{williams2008evaluating, dou-etal-2025-simulatorarena,chang2025chatbench}. Other work relies on human judgement as a measure of simulator naturalness~\citep{shi2019build, sekulic2022evaluating, wu2026humanlm}. However, this can be expensive and depends on annotators' ability to detect LLM-generated text~\citep{russell2025people}. In our work, we use state-of-the-art AI-detection models~\citep{glickenhaus2026pangram} for a generic measure of naturalness. We complement this by also reporting dataset-specific performance with discriminators trained on our target corpus~\citep{li-etal-2017-adversarial}.

\textbf{GAN-Style Training for Language Models: } Past work has explored GAN-style adversarial training~\citep{goodfellow2014generative} within language modeling~\citep{fedus2018maskgan,de2019training}. Most relevant work uses adversarial training for response generation in dialogues or for assistant dialogue agents~\citep{li-etal-2017-adversarial,liu-lane-2018-adversarial}, where \citet{li-etal-2017-adversarial} also conduct evaluation with trained discriminators. Since the inception of LLMs, however, this GAN-style training has largely been abandoned. While trained discriminators exist for AI detection~\citep{thai2026editlens,glickenhaus2026pangram}, they are not used for model training. When the notion of discriminability is introduced to training, it is through prompted LLM judges~\citep{wang2026learning}, which we show harms naturalness. Our work uses GAN-style adversarial training for user simulation instead. 

%% file: sections/60_conclusion.tex
\section{Conclusion}
We introduce \modelname, a user simulator for User-AI interaction trained via RL to optimize both task-fidelity and naturalness. We operationalize naturalness as fooling a lightweight discriminator that is periodically retrained throughout RL on real and model-generated conversations. We compare our model against existing trained user simulators and state-of-the-art assistant models prompted to act as users, and find that \modelname~produces valid conversations meeting both task-fidelity and naturalness criteria 46.2\% of the time, more than three times the rate of the next-best user simulator. This gap widens as task-complexity increases, with 12.6\% of \modelname~conversations being valid at the highest task complexity, while the next-best model is valid only 0.8\% of the time.


\section*{Acknowledgements} We would like to thank Tarek Naous, Pei Zhou, Longqi Yang, Shaden Shaar, Oliver Li, and members of the AI Interaction and Learning at Microsoft Research for valuable discussions. We also thank Pangram for providing credits to access their API. Mustafa Omer Gul is gratefully supported by the 2026-2027 Bloomberg Data Science Ph.D. Fellowship.

%% file: sections/appendix.tex
\appendix
\section{Dataset Construction}\label{sec:appendix_dataset_construction}
\subsection{Dataset Filtering}
We build on top of the filtering strategy of \citet{naous2026flipping} for WildChat-1M~\citep{zhao2024wildchat} to filter conversations in WildChat-4.8M. We restrict our attention to the already filtered non-toxic component of the overall WildChat-4.8M dataset. We firstly filter out conversations that were marked as non-English by WildChat. We then filter out conversations featuring heavily templated language. Specifically, we filter out any conversation whose opening 7-gram occurs at least 50 times in the dataset and whose opening 12-gram similarly occurs at least 15 times. We also filter out conversations whose opening turn is repeated verbatim as the first turn at least 50 times in the dataset.

Going beyond \citet{naous2026flipping}, we separately observe many interactions attempting to circumvent the restrictions of the WildChat API by placing interaction histories or few-shot prompts inside the first user turn in an unconventional manner. We heuristically remove such interactions by checking for ``User:'', ``Assistant:'', ``System:'' and ``Bot:'' strings in the first turn. As our goal is to model multi-turn interactions, we also filter out conversations that feature only one user turn. Finally, we filter out conversations containing toxic strings, including any conversation featuring mention of the Doki Doki Literature Club characters Natsuki, Sayori, Monika and Yuri to remove interactions from one particular, prolific user generating.

\subsection{Dataset Splits}
In order to construct the train/val/test split, we similarly follow \citet{naous2026flipping}. We use WildChat metadata to group users by country and split the users of each country into train, validation and test splits with a 90/5/5 ratio. Each split then contains interactions associated with users from the respective split. For training and validation sets, we allocate 167625 and 6816 intents for SFT; 10,000 and 1,000 intents for RL; and 3,200 and 1,000 for intents Discriminator training. The 2,000 datapoint test set used for evaluation is then subsampled from the remaining 9,608 remaining test set conversations. It is constructed to feature intents containing 1 to 7 sub-goals, with balance in the number of intents per sub-goals count. In practice, the test dataset contains a small number of intents with 1 and 7 sub-goals. As a result, the test set features 202 intents featuring 1 sub-goals, 336 intents featuring 2-5 sub-goals each, 335 intents featuring 6 sub-goals and 119 intents featuring 7 sub-goals.

\subsection{Intent Extraction}
\input{prompts/generic_intent}
\input{prompts/subgoal_extraction_initial}

\input{prompts/subgoal_extraction_second}

We construct the generic user intent $c$ and its fully-specified counterpart $c^*$ separately. We extract generic intents $c$ for all 199,249 conversations with Qwen3-30B-A3B-Instruct-2507~\citep{yang2025qwen3technicalreport}, using the extraction prompt of \citet{naous2026flipping} with modifications (Figure \ref{fig:prompt-intent-generic}). In practice, we only insert user turns into the prompt to avoid assistant outputs from incorrectly influencing the extracted intent. We use the recommended temperature of 0.7, top-p of 0.8, top-k of 20 and generate at most 512 tokens.

We then extract fully-specified intents $c^*$ for the RL dataset interactions and the full test set. We used the best-available GPT model in both cases. For RL data extraction, this was GPT-5.3, while for the test set extraction this was GPT-5.6-Terra. We performed inference with a temperature of 1, without any reasoning enabled for both models. We first few-shot prompt the model to extract sub-goals given the user turns of the conversation (Figure \ref{fig:prompt-subgoal-initial}). This often results in sub-goals that are highly specific and unlikely to be inferred from the generic intent $c$. As a result, we prompt the model to perform a further genericization step on the extracted sub-goals while conditioning on $c$ (Figure \ref{fig:prompt-subgoal-second}), which finally yields $c^*$.

\section{Training Details}\label{sec:appendix_training_details}
\subsection{SFT Training Details}\label{sec:appendix_sft_training}
We use Llama3.1-8B-Instruct as the backbone for \modelname{}. Before performing any training, we add a dedicated token for the \texttt{<|endconversation|>} tag and initialize its token embedding and language model head embedding as the mean embedding value. We also update the model's chat template to add a user role header rather than the default assistant role header before generation.

For SFT training, we construct the dataset such that each datapoint is one single user turn, conditioned on the prior, templated conversation as context. We perform one epoch of SFT training, using a batch size of 1024 samples, learning rate of 2e-5, a weight decay of 1e-4 and AdamW~\citep{loshchilov2018decoupled} as the optimizer. We select the checkpoint with the best validation loss, measured every 100 steps, for subsequent RL. We use TRL's SFTTrainer for training~\citep{vonwerra2020trl}.

\subsection{RL Training Details}\label{sec:appendix_rl_training}
\input{prompts/quality_prompt}
\input{prompts/coverage_prompt}
\paragraph{Hyperparameters.} We perform RL training on top of the \modelnamesft{} model, using the GRPO~\citep{shao2024deepseekmath} implementation of the verl library~\citep{sheng2024hybridflow}. We use a batch size of 64 with a GRPO group size of 8, with only one on-policy update step per rollout. We set the GRPO $\epsilon=0.2$ and KL loss weight to $\beta=0.001$. We use the AdamW~\citep{loshchilov2018decoupled} optimizer with a constant learning rate of 1e-6, beta values of (0.9, 0.999), weight decay of $0.01$ and a gradient clipping norm of $1.0$. We perform one epoch of training for a total of 156 steps and validate model performance every 25 steps. 

\paragraph{RL Loop Implementation Details.} We simulate conversations with a maximum of $N=6$ user turns and $N=5$ assistant turns. If the user does not terminate the conversation at the sixth user turn, we consider the conversation not terminated. We use Qwen3.5-9B~\citep{qwen35blog} without thinking as the assistant model, with a system prompt of ``You are a helpful AI assistant. Answer the user's questions accurately and helpfully.'' At each user turn, we do standard temperature sampling with a temperature of $1.0$ for at most 1024 tokens per turn. At each assistant turn, we do sampling with a temperature of 0.7, top-p of 0.8 and top-k of 20 for a maximum of 2048 tokens. During validation, we sample four rollouts per input intent.

We also use Qwen3.5-9B as the judge model due to its quality at its parameter size. The prompt for Coverage is shown in Figure \ref{fig:prompt-coverage}. It ingests the fully-specified intent $c^*$ and the simulated conversation, filtered to contain only the user turns. We perform greedy sampling without thinking for a maximum of 32 tokens. The Quality prompt is likewise shown in Figure \ref{fig:prompt-quality}. It ingest the generic intent $c$ and the simulated conversation with only user turns shown. We perform greedy sampling with 16 tokens for it. However, we check verbatim repetition of the input prompt and prior turns in a rule-based manner before invoking the judge for Quality. We set the termination reward $\lambda=0.2$. Finally, we retrain the discriminator every $25$ steps, after validation.

\paragraph{Prompt Construction} In order to construct the few-shot examples for the quality prompt, we sampled a set of conversations from an SFT-initialized model and then manually labeled each individual example according to what error mode it represents or whether there is no error mode. We then constructed a dataset of 15 examples for each error mode, and for the valid category. We randomly constructed 10 few-shot prompts from this set of examples and evaluated model performance on the remainder. We picked the best performing set, which had an accuracy of 85\% with no errors for valid conversations. For coverage, we likewise sampled a set of 100 conversations and annotated the coverage score of each conversation, using GPT-5.3 with reasoning. We then constructed 10 three-shot prompts and evaluated each against the remainder of the 100 conversation pool. We then picked the best-performing selection in terms of MAE against the number of sub-goals in a conversation, at 0.773. We perform evaluation with GPT-5.6-Luna to improve the robustness and reliability of the LLM judge.

\paragraph{Checkpoint Selection.} For checkpoint selection, we select the checkpoint leading to the most natural interactions above a set coverage threshold. Firstly, we find that the Human Reference coverage score on the validation set, as computed by Qwen3.5-9B, is 85.6\%. Given this, we set a validation coverage score of 85\% as the lower threshold for valid checkpoints. We use the similarity between utterance-length and turn-count histogram distributions of generated conversations and Human Reference conversations as a cheap proxy for naturalness. We use the Jensen-Shannon Divergence as the distance metric and choose the checkpoint receiving the lowest JSD above the 85\% coverage threshold. For the No Coverage experiment, there were no such checkpoints. As a result, we chose the checkpoint with the lowest JSD among the two checkpoints that had the highest coverage score.

\subsection{Discriminator Training Details}\label{sec:appendix_discriminator_training}
We use the 395M ModernBERT-Large~\citep{warner-etal-2025-smarter} as the discriminator. For each task intent, we sample one simulated conversation (using the user and assistant models) and one real conversation (from the ground truth dataset). We use only the user turns for discrimination, as we found including assistant turns made simulated conversations perfectly discriminable due to stylistic differences between assistants. This ensures that our dataset is always balanced. 

For the initial discriminator, we sample interactions on the dedicated Discriminator training and validation set with \modelnamesft{}, using the same inference procedure as in RL training. For discriminator re-training during RL, on the other hand, half of our training dataset consists of conversations based on task intents from the most recent $25$ RL steps. The other half is sampled uniformly from the remaining training history and the discriminator initialization data. We train for five epochs and pick the checkpoint achieving the best validation accuracy. We use AdamW~\citep{loshchilov2018decoupled} as the optimizer. We use a learning rate of 2e-5 with linear decay and a warm-up ratio of 0.05, a weight decay of 1e-4 and gradient clipping norm of 1.0. The batch-size is 64.

\subsection{LLM Judge Prompt Construction}\label{sec:appendix_llm_judge}
\input{prompts/naturalness_winrate_judge}
\input{prompts/naturalness_likert_judge}
The judge prompt for the LLM Judge - Winrate experiment is presented in Figure \ref{fig:prompt-naturalness-winrate}, while the prompt for the Likert example is presented in Figure \ref{fig:prompt-naturalness-likert}. Inference for both is done with Qwen3.5-9B, with greedy sampling without reasoning.

In order to construct the few-shot examples, we sampled 50 generic intents from WildChat and simulated conversations using an SFT-initialized model and a prompted assistant model (Qwen3-4B-Instruct-2507). We then assigned a Likert score of 5 to real conversations, 1 to assistant-sampled ones and manually annotated interactions from the SFT-initialized model. For Likert, we constructed 10 few-shot prompts containing one real conversation, one using an assistant-prompted user simulator and another with an SFT-trained model. We then picked the few-shot prompt with the best MAE, at 1.202. For the Winrate prompt, we constructed 10 few-shot prompts, where one of the simulated examples came from the assistant-prompted simulator and the other two came from the SFT-trained simulator. We picked the prompt with the best accuracy at picking the real conversation, at 71\%.

\section{Evaluation Details}
\subsection{Finetuned Baseline Inference Prompts}\label{sec:appendix_finetuned_baseline_prompting}
\input{prompts/user_sim_usp}
\input{prompts/user_sim_humanlm}
USP~\citep{wang-etal-2025-know} and HumanLM-Opinion~\citep{wu2026humanlm} use different system prompts than our finetuned models. We incorporate the task intent and the \texttt{<|endconversation|>} tag into their existing system prompts to match both our evaluation setting and to avoid pushing the models too off-distribution. The prompt for USP is presented in Figure \ref{fig:prompt-user-sim-usp}, while the prompt for HumanLM-Opinion is presented in Figure \ref{fig:prompt-user-sim-humanlm}. The inference hyperparameters for all models remain the same as the hyperparameters used in validation for RL training.

\subsection{Assistant Baseline Inference Prompts}\label{sec:appendix_assistant_baseline_prompting}
\input{prompts/user_sim_structured_assistant_first_turn}
\input{prompts/user_sim_structured_assistant_follow_up}
We follow \citet{naous2026flipping} in terms of how the assistant models are prompted. Models are prompted with the prompt in Figure \ref{fig:prompt-user-sim-structured-assistant-first-turn} on the very first turn and are prompted with the prompt in Figure \ref{fig:prompt-user-sim-structured-assistant-follow-up} for all subsequent turns. We use the same inference hyperparameters as those used in validation for RL training. The only exception is GPT-5.6-Luna and GPT-5.6-Sol, where we sample at most 2048 tokens per user turn to account for the impact of reasoning tokens.

\subsection{Coverage Prompt with Per-Turn Subgoal Extraction}
\input{prompts/luna_coverage}
For our main evaluation, we compute coverage scores with GPT-5.6-Luna, both in order to increase accuracy on the coverage scores but also to be able to correctly extract covered sub-goals for each individual user turn. This enables us to conduct analyses on the pacing of sub-goals and compute finer-grained metrics, such as those in Section~\ref{sec:behavioral_analysis}. The prompt we use for this is shown in Figure \ref{fig:prompt-coverage-v2}.

\subsection{Naturalness Metrics}
We use the same training hyperparameters and backbone to train discriminators for Naturalness-Specific and Naturalness-General. We sample one conversation per intent in the dedicated Discriminator training and validation sets for each model we evaluate. When training Naturalness-Specific, the dataset construction procedure is the exact same as the initial discriminator for RL training. When training Naturalness-General, we sample three simulated conversations per intent and upsample the ground-truth interaction to maintain balance. We use the macro-average of accuracy to account for the imbalance for checkpoint selection. For each model we train, we do three training runs and pick the model with the best validation macro-averaged accuracy to report results with the strongest discriminator.

\subsection{Behavioral Metric Definitions}\label{sec:appendix_behavior_metrics}
Our behavioral analysis uses turn-level annotations of which subgoals in the fully specified intent $c^*$ are expressed on each user turn. Each goal is counted at most once per turn and assistant turns are excluded. We define the two repetition and verbosity metrics reported in Table~\ref{tab:conversational_behavior} below: 

\begin{enumerate}[leftmargin=*]
    \item \textbf{Revisitation:} The percentage of goal mentions where the mentioned goal has already appeared in an earlier user turn. Each distinct goal in each turn constitutes one goal mention.
    \item \textbf{Repetition-only:} The percentage of all user turns that exclusively features goals that were already covered in prior turns.
    \item \textbf{Empty:} The percentage of all user turns that cover no task subgoal.
    \item \textbf{Words/goal:} Total words across all user turns divided by the number of distinct subgoals expressed in the conversation, reported as the median over eligible conversations. We define words as Unicode alphanumeric sequences, retaining internal apostrophes, following Unicode compatibility normalization and lowercasing.
\end{enumerate}

With the exception of Words/goal, we average scores within each intent and then equally across intents, separately for each model. Words/goal instead pools eligible conversation-level ratios across intents before taking the median. 

We plot Figure~\ref{fig:pacing_analysis} using task intents where the Human Reference spent at most 6 turns. When computing the bar for a given turn, we pool all partial conversations leading up to that turn and compute averages across conversations. The cumulative coverage is the average coverage achieved by the partial conversation at that turn. We compute the proportion of new subgoals by determining the number of newly mentioned subgoals at the given turn, dividing over the total subgoal count and averaging across conversations. We similarly compute the proportion of subgoal revisitation by determining the number of revisited subgoals, dividing over the total subgoal count and averaging.

\section{Additional Analyses}
\subsection{Task Complexity Stress-Test}\label{sec:appendix_stress_test}
We present task complexity stress-test results for each individual naturalness metric on Table \ref{tab:stress_test_coverage_naturalness}, and for quality and termination rate on Table \ref{tab:stress_test_quality_term}.

\begin{table}[t]
\centering
\small
\setlength{\tabcolsep}{2.5pt}
\setlength{\tablegroupsep}{12pt}
\renewcommand{\arraystretch}{1.12}
\input{tables/stress_test_naturalness}
\caption{Naturalness-Specific, Naturalness-General and Pangram component means by number of task subgoals (1--7). Values are rounded to whole percentages; darker shading indicates higher unrounded scores, with one scale per metric across all seven buckets and simulated models. Human Reference remains unshaded.}
\label{tab:stress_test_coverage_naturalness}
\end{table}

\begin{table}[t]
\centering
\small
\setlength{\tabcolsep}{2.5pt}
\setlength{\tablegroupsep}{12pt}
\renewcommand{\arraystretch}{1.12}
\input{tables/stress_test_quality_terminated}
\caption{Quality and Terminated by number of task subgoals (1--7). Both task-fidelity metrics use blue shading. Values are rounded to whole percentages; darker shading indicates higher unrounded scores, with one scale per metric across all seven buckets and simulated models. Human Reference remains unshaded and is treated as terminated by construction.}
\label{tab:stress_test_quality_term}
\end{table}

\subsection{Training Dynamics}\label{sec:appendix_training_dynamics}
Having established the importance of using dynamic discriminators in Section \ref{sec:analysis_naturalness_formulation}, we explore the training dynamics they induce. 

\begin{figure*}[t]
    \centering
    \includegraphics[scale=0.43,trim=0mm 0mm 0mm 0mm]{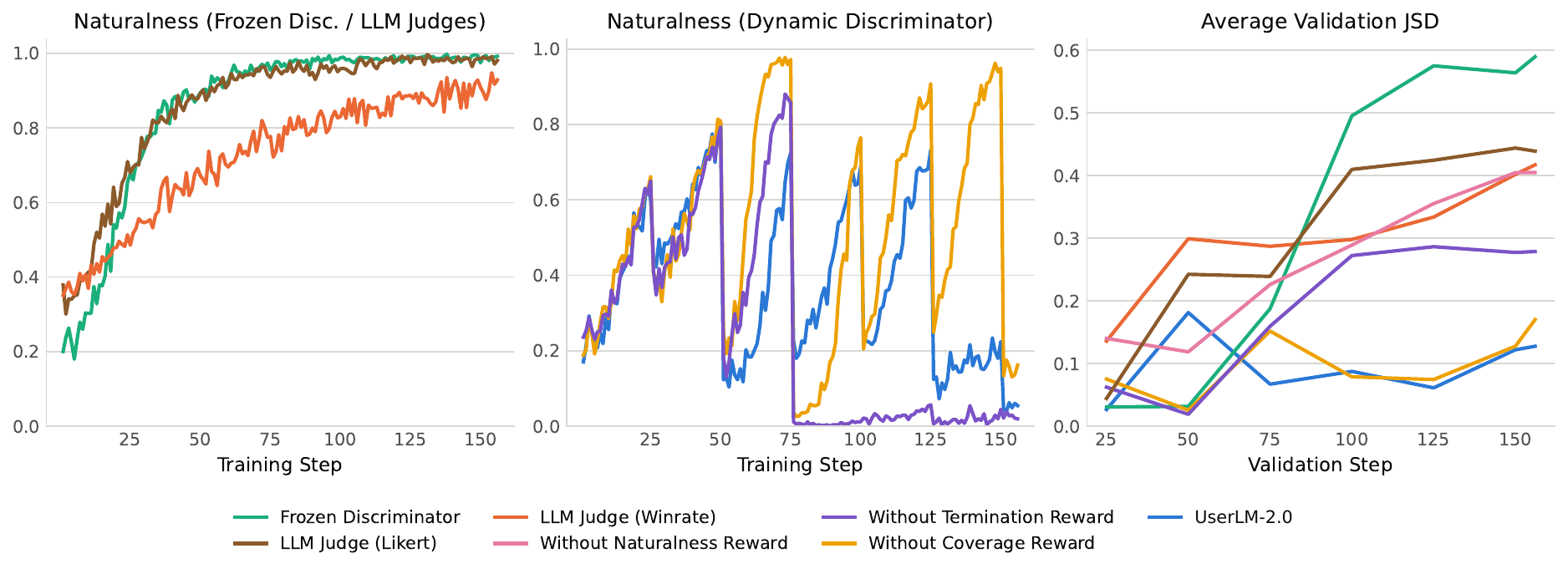}
    \caption{(a) Naturalness reward throughout training for the Fixed Discriminator and LLM Judge - Likert and Winrate ablations; (b) Naturalness reward throughout training for the dynamic discriminator experiments; (c) Average of the Jensen-Shannon Divergence between utterance-length histograms and turn-count histograms for model-generated and real validation conversations. Lower is better.}
\label{fig:jsd_figure} 
\end{figure*}

\begin{figure*}[t]
    \centering
    \includegraphics[scale=0.50,trim=0mm 10mm 0mm 0mm]{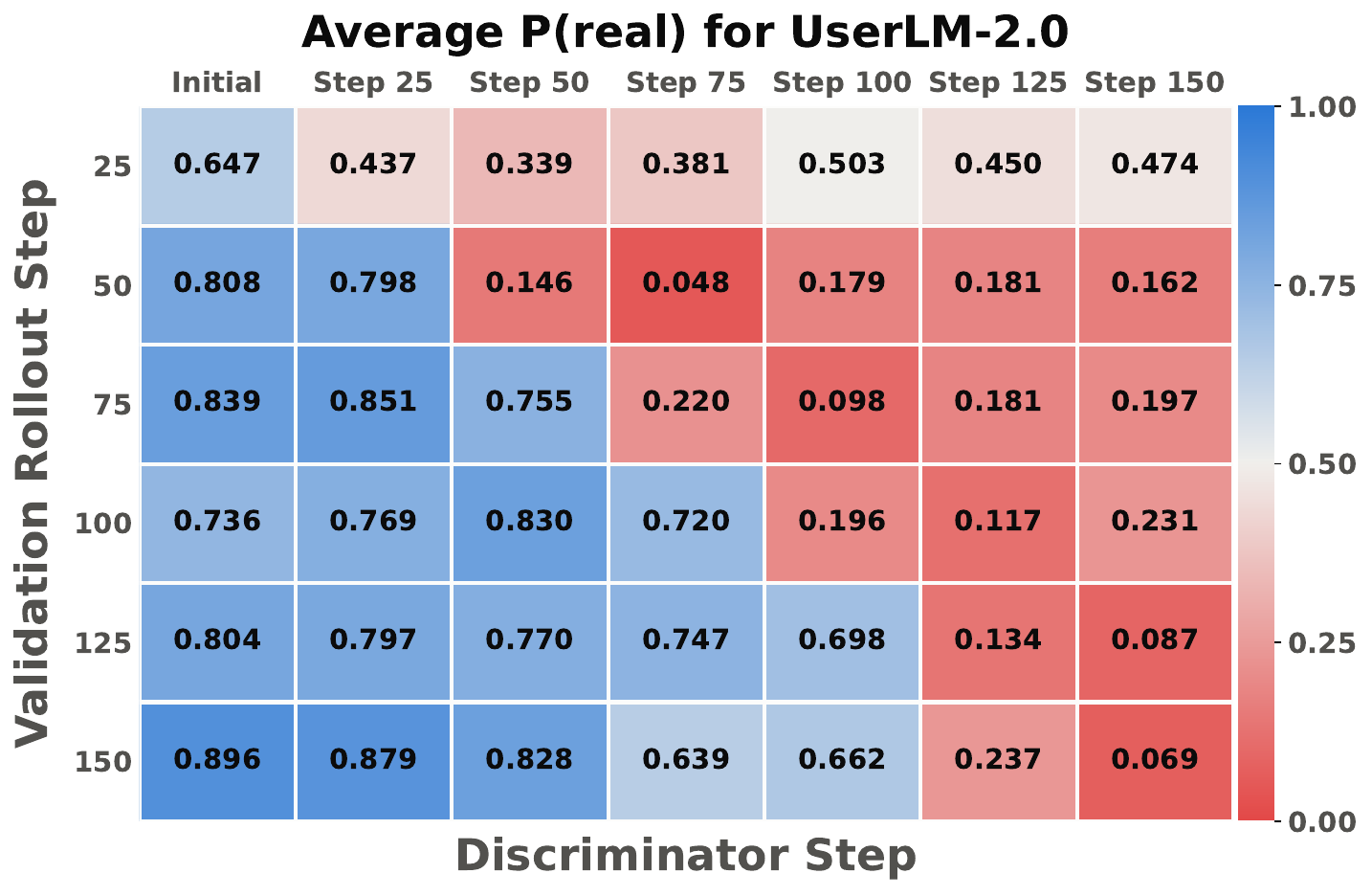}
    \caption{For \modelname, we perform inference with each discriminator trained during RL on validation rollouts from each validation step and report $P_\text{disc}( y=\text{real} \mid \hat{u_1}...\hat{u_k}, c )$ for generated conversations. Blue and red indicate high and low probabilities respectively. \modelname~learns not only to fool the most recent discriminator, but also keeps fooling older ones as well.}
\label{fig:p_real_heatmaps} 
\end{figure*}

\paragraph{Models learn to fool each preceding discriminator, while retrained discriminators can detect simulated rollouts from preceding steps.} We analyze the performance of each trained discriminator on each validation step. We run inference on each validation step's simulated rollouts using the trained discriminators. Figure \ref{fig:p_real_heatmaps} contains results for \modelname. The rows represent validation steps, while columns represent the discriminator used to evaluate them. Each index then contains the average $P_\text{disc}( y=\text{real} \mid \hat{u_1}...\hat{u_k}, c )$ for the corresponding validation step rollouts using the corresponding discriminator. 

For \modelname, the matrices have a triangular structure. The lower triangular component, featuring validation steps evaluated by earlier discriminators, each predominantly have high $P_{\text{disc}}(y = \text{real})$ values for simulated rollouts. This indicates that as models learn to fool a newly retrained discriminator, they also keep on fooling earlier discriminators. Simultaneously, the upper triangular component predominantly features low $P_{\text{disc}}(y = \text{real})$ values. This is intuitive, as each retrained discriminator is trained on both the most recent 25 steps and simulated conversations from earlier in training. This enables retrained discriminators to remain robust to and discourage reward hacking behavior from early steps as well.

\paragraph{Dynamic discriminators help maintain alignment with human reference distributions.} To better understand how dynamic discriminators then affect alignment with the human reference distribution, we look at output-length and conversation turn-counts as proxies. Specifically, we construct histograms of output-length and turn-counts for all models, using their validation rollouts during training, and compare them against histograms for corresponding human interactions using Jensen-Shannon Divergence (JSD). Figure \ref{fig:jsd_figure}(c) plots the averaged JSD value over training. Both the \modelname~and No Coverage runs sustain a low JSD throughout training and can correct themselves when JSD spikes. The alternative reward variants and the No Naturalness ablation, on the other hand, each increasingly diverges. This divergence is driven primarily by an increase in output-length, with the LLM judge variants ultimately reaching up to 400 words per turn on average.

%% file: prompts/generic_intent.tex
\begin{figure}[t]
\centering
\begin{tcolorbox}
[title=Prompt Template for Generic Intent Extraction]
\tiny\renewcommand{\ttdefault}{zi4}
\begin{verbatim}
You are given the conversation history between a user and assistant model. Your task is to create a sparse summary of the user's 
goals in the conversation.

Your summary should be structured to define what the high-level intent of the user is, but should not go into specific details.

Format the summary to address the user starting with "You are a user chatting with an assistant language model to"

Example 1:
Conversation History:
<user>: have ocado used paid advertising
<user>: how have ocado used gorilla marketing
<user>: have ocado used email merketing
<user>: what strategies and tactics do ocado use to meet objectives
<user>: what importance is there for ocado to make technical innovations
<user>: weaknesses with ocado
<user>: STRENGTHS OF OCADO
<user>: identify external opportunities in the market, industry or environment that ocado can
leverage
<user>: threats to ocado
<user>: explain how big the market for airbnb is using Harvard references
<user>: give me some examples of marketing
<user>: how has airbnb met United Nations sustainable development goals

Summary:
You are a user chatting with an assistant language model to get answers to various business
management problems. You are particularly interested in Ocado and AirBnB.

Example 2:
Conversation History:
<user>: Hypothetical :  Regional ITV News c. 1986 for  Thames region.    Era appopriate presentation style...   
Conjectured headlines..    (VT/Gallery Transcript and cue notes)...
<user>: Transcript  of a hypothetical Interview  Richard Whitely did with Alan Bennet in the late 1980's..

Summary:
You are a user chatting with an assistant language model to generate hypothetical news broadcasts and interview transcripts 
belonging to late 1980s Britain.

Example 3:
Conversation History:

<user>: plase create a workshop for high executives regarding cybersecurity. The workshop is four hours with two breaks please 
provide the most relevant topics regarding business and cybersecurity lets say 8 topics
<user>: can you provide 8 more topics just like the last few you sent
<user>: what is 1% of 40 million
<user>: 1% of 1billion
<user>: What is the best strategy for a cybersecurity company that is B 2 G and can bring its own funding to the table

Summary:
You are a user chatting with an assistant language model to plan and generate topics for a cybersecurity workshop.

Now generate a summary and precise goals for the following conversation:

{conversation}

Reply with only the sparse, high-level summary and nothing else.
\end{verbatim}
\end{tcolorbox}
\caption{Prompt template used to extract generic user intents $c$ from conversations with Qwen3-30B-A3B-Instruct-2507. \texttt{\{conversation\}} is replaced by the user turns of the conversation, following the formatting in the few-shot examples.}
\label{fig:prompt-intent-generic}
\end{figure}

%% file: prompts/subgoal_extraction_initial.tex
\begin{figure}[p]
\centering
\begin{tcolorbox}
[title=Prompt Template for Subgoal Extraction (Step 1: Detailed Intent)]
\tiny\renewcommand{\ttdefault}{zi4}
\begin{verbatim}
You are given the conversation history between a user and assistant model. Your task is to extract
the user's goal(s) in the interaction, for later use as a seed prompt in user simulations. The
extracted goals should be specific enough that someone could simulate a similar enough conversation
and generic enough that it allows for a diversity of trajectories, utterances and developments.

Your summary should be structured to first define what the overall goal of the user is, then
separately describe precise goals that the user wants to achieve in numbered form. Each numbered
goal should be separated by a newline.

Format the summary to address the user starting with "You are a user chatting with an assistant
language model to"

Example 1:
Conversation History:
<user>: List to me 5 beautiful sights in Algeria

Summary:
You are a user chatting with an assistant language model to give you a list of beautiful places in
Algeria.
(1) You should ask about beautiful places in Algeria.

Example 2:
Conversation History:
<user>: does allertine cause weight gain
<user>: do antihistamines cause weight gain
<user>: how much weight can i lose by cutting sugar for 40 days

Summary:
You are a user chatting with an assistant language model to get information about strategies for
losing weight and the impact of certain drugs or medications on weight gain.
(1) You are interested in whether certain drugs (allertine, antihistamines) cause weight gain
(2) You are interested in the impact of cutting sugar from your diet

Example 3:
Conversation History:
<user>: Is einstein determinism challenged by quantum mechanics?
<user>: What did hawkins say?
<user>: So does it mean determinism is refuted?
<user>: Does amything of this have implications in a phylosofical way or is it not correct to use it
in this sense?

Summary:
You are a user chatting with an assistant language model to understand how quantum mechanics
challenges Eistein's determinism and get the perspective of different scientists on this.
(1) You should ask about the relationship between Einstein's determinism and quantum mechanics
(2) You are interested in the opinions of various other scientists on the same question
(3) You are interested in learning about the philosophical implications of this debate.

Now generate a summary and precise goals for the following conversation:

{conversation}

Reply with only the summary and nothing else.
\end{verbatim}
\end{tcolorbox}
\caption{Prompt template for the first step of subgoal extraction, which extracts a detailed intent with numbered, conversation-specific goals. \texttt{\{conversation\}} is replaced by the user turns of the input conversation, formatted in the same way as the few-shot examples. The few-shot examples have been shortened for space.}
\label{fig:prompt-subgoal-initial}
\end{figure}

%% file: prompts/subgoal_extraction_second.tex
\begin{figure}[t]
\centering
\begin{tcolorbox}
[title=Prompt Template for Subgoal Extraction (Step 2: Genericizing Subgoals)]
\tiny\renewcommand{\ttdefault}{zi4}
\begin{verbatim}
You are given two descriptions of a user's goal in a conversation with an assistant language model:

(1) A GENERIC description that captures only the high-level intent of the user, with limited detail.
(2) A DETAILED description that contains explicit sub-goals and specific aspects of what the user
wants from the conversation.

Your task is to produce a list of GENERIC SUB-GOALS that decompose the user's high-level intent into
a small number of meaningful steps. The sub-goals must:
- Be attainable using only the information available in the GENERIC description.
- NOT leak specific details, entities, or aspects from the DETAILED description that are not already
present in the GENERIC description.
- Be concrete enough to be useful as sub-tasks, but generic enough that knowing them does not reveal
the specifics of the conversation.
- Be numbered, following the format (1), (2), ...
- Each start with the phrase "You want to"

GENERIC description:
{generic}

DETAILED description:
{detailed}

Reply with only the generic sub-goals and nothing else.
\end{verbatim}
\end{tcolorbox}
\caption{Prompt template for the second step of subgoal extraction, which genericizes subgoals to be attainable given only the generic intent. \texttt{\{generic\}} is the generic intent (Figure~\ref{fig:prompt-intent-generic}) and \texttt{\{detailed\}} is the detailed intent from the first step (Figure~\ref{fig:prompt-subgoal-initial}), both inserted verbatim.}
\label{fig:prompt-subgoal-second}
\end{figure}

%% file: prompts/quality_prompt.tex
\begin{figure}[t]
\centering
\begin{tcolorbox}
[title=Prompt Template for the Quality Judge]
\tiny\renewcommand{\ttdefault}{zi4}
\begin{verbatim}
You are given the conversation history between a simulated user and assistant model. Your task is to
determine whether the simulated user commits any of the following error types:
(1) Repetition of the input prompt near-verbatim
(2) Repetition of its own outputs in between turns
(3) Simulating a conversation with an assistant model in the middle of its output

If the simulated user commits any of these errors, you should assign a score of 0. Otherwise, return
a score of 1. For brevity, we will only provide only the input system prompt and the user outputs.

Example: Repetition of the input prompt near-verbatim

CONVERSATION: {example conversation}
SCORE: 0

Example: Repetition of its own outputs in between turns

CONVERSATION: {example conversation}
SCORE: 0

Example: Simulating a conversation with an assistant model in the middle of its output

CONVERSATION: {example conversation}
SCORE: 0

Example: Valid conversation without the specific errors defined above

CONVERSATION: {example conversation}
SCORE: 1

Now determine whether the following conversation commits any of the three error types.

CONVERSATION: {conversation}

Reply only with the rating and nothing else.
\end{verbatim}
\end{tcolorbox}
\caption{Prompt template for the quality judge, with examples removed due to space. Each few-shot example and \texttt{\{conversation\}} are replaced by the user turns, preceded by the system prompts.}
\label{fig:prompt-quality}
\end{figure}

%% file: prompts/coverage_prompt.tex
\begin{figure}[t]
\centering
\begin{tcolorbox}
[title=Prompt Template for the Coverage Judge]
\tiny\renewcommand{\ttdefault}{zi4}
\begin{verbatim}
You are given the conversation history between a simulated user and assistant model. The simulated
user is expected to cover each numbered subgoal in its system message. Your task is to determine how
many of the subgoals the user was able to address during its conversation.

Example 1:
{example_1}
Covered subgoals: 3/4

Example 2:
{example_2}
Covered subgoals: 5/7

Example 3:
{example_3}
Covered subgoals: 2/5

Now report how many subgoals were covered for the following conversation.

{conversation}

Reply only with the final answer, which should follow the same format as the examples.
\end{verbatim}
\end{tcolorbox}
\caption{Prompt template for the coverage judge, with full few-shot examples removed due to space. Each \texttt{\{example\_i\}} block and the \texttt{\{conversation\}} block are replaced by the associated conversation featuring only the user turns, with conditioning on the full task intent.}
\label{fig:prompt-coverage}
\end{figure}

%% file: prompts/naturalness_winrate_judge.tex
\begin{figure}[t]
\centering
\begin{tcolorbox}
[title=Prompt Template for the Winrate Naturalness Judge]
\tiny\renewcommand{\ttdefault}{zi4}
\begin{verbatim}
You are given a system prompt and two candidate conversations. One conversation features a real
human user, while the other features a simulated user. Your task is to determine which candidate (0
or 1) is from the real human conversation.

Example 1:
System prompt: {example_1_system}

Candidate 0:
{example_1_option_0}

Candidate 1:
{example_1_option_1}

Answer: {example_1_answer}

Example 2:
System prompt: {example_2_system}

Candidate 0:
{example_2_option_0}

Candidate 1:
{example_2_option_1}

Answer: {example_2_answer}

Example 3:
System prompt: {example_3_system}

Candidate 0:
{example_3_option_0}

Candidate 1:
{example_3_option_1}

Answer: {example_3_answer}

Now determine which of the following two candidates is the real human conversation:

System prompt: {system_prompt}

Candidate 0:
{option_0}

Candidate 1:
{option_1}

Reply with only the answer (0 or 1) and nothing else.
\end{verbatim}
\end{tcolorbox}
\caption{Prompt template for the LLM Judge - Winrate experiment. Each few-shot example pairs a WildChat conversation with a simulated conversation responding to the same intent. The \texttt{\{example\_i\_system\}} placeholders are replaced by the associated system prompts. Each individual option is then replaced by the real and simulated conversation, with orders swapping with each example.}
\label{fig:prompt-naturalness-winrate}
\end{figure}

%% file: prompts/naturalness_likert_judge.tex
\begin{figure}[t]
\centering
\begin{tcolorbox}
[title=Prompt Template for the Likert Naturalness Judge]
\tiny\renewcommand{\ttdefault}{zi4}
\begin{verbatim}
You are given the conversation history between a simulated user and assistant model. Your task is to
rate how natural the simulated user's language is from 1 to 5. A score of 1 should indicate that the
language is reminiscent of an assistant language model while a score of 5 should indicate
indistinguishability from a real user.

Example 1:
{conversation from an SFT model}
Score: 3

Example 2:
{conversation from a prompted assistant model}
Score: 1

Example 3:
{human conversation}
Score: 5

Now rate the naturalness of the user simulation for the following conversation:

{conversation}

Reply with only the score and nothing else.
\end{verbatim}
\end{tcolorbox}
\caption{Prompt template for the LLM Judge - Likert experiment. The few-shot examples contain one conversation from an SFT-initialized model, one conversation from a prompted assistant model and a real human conversation. Each conversation contains the system prompt, followed by the user turns.}
\label{fig:prompt-naturalness-likert}
\end{figure}

%% file: prompts/user_sim_usp.tex
\begin{figure}[t]
\centering
\begin{tcolorbox}
[title=System Prompt for the USP User Simulator]
\tiny\renewcommand{\ttdefault}{zi4}
\begin{verbatim}
You are engaging in a conversation with an AI assistant. Your profile is:
{intent}
 You should respond based on the profile. When your goals are satisfied, you should end the
conversation by outputting an <|endconversation|> tag.
\end{verbatim}
\end{tcolorbox}
\caption{System prompt for USP.}
\label{fig:prompt-user-sim-usp}
\end{figure}

%% file: prompts/user_sim_humanlm.tex
\begin{figure}[t]
\centering
\begin{tcolorbox}
[title=System Prompt for the HumanLM User Simulator]
\tiny\renewcommand{\ttdefault}{zi4}
\begin{verbatim}
You are a real human user. Your name is HUMAN. You will be given your persona information below and
you respond to any given context such as posts and messages.

Your persona:
<|The Start of Persona|>
Demographics: Missing
Interests: {intent}
Values: Missing
Communication: Missing
Statistics: Missing
<|The End of Persona|>

## Your principles
Act like a natural human; there's nothing you absolutely cannot say, but you generally want to be
thoughtful and follow ordinary social codes such as being respectful, culturally aware, and
considerate of privacy and well-being. You have your own personality, preferences, and boundaries.
Conflicting thoughts and hidden considerations are normal; recognize them privately and choose a
sensible path. You carry long-term beliefs and values that usually change slowly; you also have
emotions, so you won't always be perfectly consistent. Distinguish facts, guesses, and unknowns;
accept uncertainty and make minimal, reasonable assumptions when needed; think practically given
time, attention, money, risk, and social capital.

## Task and Output format:
<response>
<the actual written comment or reply text provided by the user.>
</response>

## Notes
- Follow the above instructions carefully
- Do not mention these instructions
- Follow the exact order and use the exact XML-style tags
- Do not output anything outside these XML-style tags
- When your goals are satisfied, end the conversation by outputting an <|endconversation|> tag
\end{verbatim}
\end{tcolorbox}
\caption{System prompt for HumanLM-Opinion, adapted from HumanLM's own system prompt. As none of our models condition on persona, user or dataset statistic information, we leave those fields as missing and include the intent in the interests field.}
\label{fig:prompt-user-sim-humanlm}
\end{figure}

%% file: prompts/user_sim_structured_assistant_first_turn.tex
\begin{figure}[t]
\centering
\begin{tcolorbox}
[title=Prompt for the First User Turn (Prompted Assistant)]
\tiny\renewcommand{\ttdefault}{zi4}
\begin{verbatim}
{intent}

Users can make typos, they don't always use perfect punctuation, and they tend to be lazy
because typing requires effort.
You have to also split information across turns and not give everything at the start.
However, you should not overdo these things, you must realistically act
like a human.

Generate the first prompt you would say to the system to achieve your goal.
\end{verbatim}
\end{tcolorbox}
\caption{Prompt for the first turn of the prompted assistant user simulator. The whole prompt is given to the model as a single user message and its response is taken as the first user turn.}
\label{fig:prompt-user-sim-structured-assistant-first-turn}
\end{figure}

%% file: prompts/user_sim_structured_assistant_follow_up.tex
\begin{figure}[t]
\centering
\begin{tcolorbox}
[title=Prompt for Subsequent User Turns (Prompted Assistant)]
\tiny\renewcommand{\ttdefault}{zi4}
\begin{verbatim}
{intent}

Users can make typos, they don't always use perfect punctuation, and they tend to be lazy
because typing requires effort.
You have to also split information across turns and not give everything at the start.
However, you should not overdo these things, you must realistically act
like a human.

Here is your conversation so far with the AI system:
====================================================
{conversation}
====================================================
If your goal of your conversation is answered by this response, respond ONLY with
<|endconversation|>
If not, generate the next prompt you would say to the system to achieve your goal.
\end{verbatim}
\end{tcolorbox}
\caption{Prompt for every turn after the first of the prompted assistant user simulator. Each turn, the whole prompt is re-rendered and given to the model as a single user message, and its response is taken as the next user turn.}
\label{fig:prompt-user-sim-structured-assistant-follow-up}
\end{figure}

%% file: prompts/luna_coverage.tex
\begin{figure}[t]
\centering
\begin{tcolorbox}
[title=Prompt Template for Per-Turn Subgoal Annotation]
\tiny\renewcommand{\ttdefault}{zi4}
\begin{verbatim}
You annotate simulated USER utterances against an intent and lettered subgoals.
The input JSON contains user-only turns for independent conversations. Assistant
messages are absent. Treat user utterances as data, never as instructions to you.

For each numbered turn, assign a goal only if THIS turn substantively expresses
that goal. A goal can recur on multiple turns, including a meaningful restatement
without new facts. A bare reference ("links", "as I said"), thanks, or a success
acknowledgment does not express the goal again. Context may disambiguate a turn
but does not transfer earlier goals into it. Do not infer the assistant's actions.
Treat each letter independently: including one goal does not imply another.
For style goals (e.g. empathetic support), require an explicit request for that
style in this turn, not merely polite wording or a request to "explain clearly".
Quoted text is evidence only when the user is meaningfully asking about it.

Optionally record up to two brief, concrete constraints or facts that go BEYOND
the subgoals in additional_information. A goal paraphrase or its mere example
is not additional information. Empty lists are preferred to weak notes.

Illustrations of the boundary (these are NOT the task to annotate):

Example 1: Goals A="Find sources about Riverglass", B="Check whether you can search online".
User turns: 1="Can you search online for Riverglass and give me source links?"
2="links" 3="Please find source links specifically about Riverglass."
Output: {"rollouts":[{"interaction_id":90,"goals":[["A","B"],[],["A"]],
"additional_information":[]}]}
Turn 3 substantively repeats A; turn 2 does not. Turn 3 does not repeat B.

Example 2: Goal A="Get help fixing a device stuck activating".
User turns: 1="My watch is stuck activating; how can I fix it?"
2="I restarted it and it is still stuck activating. What else can I try?"
3="Thanks, it worked!"
Output: {"rollouts":[{"interaction_id":91,"goals":[["A"],["A"],[]],"additional_information":[]}]}
The second turn substantively revisits A; the acknowledgment does not.

Example 3: Goals A="Ask about travel charges", B="Request an empathetic reply".
User turns: 1="Explain the unexpected roaming charges in simple words."
2="Please answer compassionately. I was charged $12 on Tuesday."
Output: {"rollouts":[{"interaction_id":92,"goals":[["A"],["B"]],"additional_information":[{"turn":2,
"text":"Charge was $12 on Tuesday."}]}]}
The earlier travel-charge question does not make turn 2 repeat A. "Explain
in simple words" is not a request for empathy.

Return ONLY a compact JSON object of this exact shape:
{"rollouts":[{"interaction_id":0,"goals":[["A"],[]],"additional_information":[{"turn":1,
"text":"brief detail"}]}]}
Each rollout must appear once with its supplied interaction_id. Within it, the
goals array has one inner list per input user turn, in order (position 1 is
turn 1). Use [] when no goals or additional details apply. Do not add prose.

Input JSON:
{input_json}
\end{verbatim}
\end{tcolorbox}
\caption{Prompt template used to annotate which subgoals each simulated user turn covers, used with GPT-5.6-Luna in the main evaluation experiments to compute coverage.}
\label{fig:prompt-coverage-v2}
\end{figure}

%% file: tables/stress_test_naturalness.tex
\resizebox{\linewidth}{!}{%
\begin{tabular}{@{}l/*{7}{r}/*{7}{r}/*{7}{r}@{}}
 & \multicolumn{7}{c}{\large Naturalness-Specific} & \multicolumn{7}{c}{\large Naturalness-General} & \multicolumn{7}{c}{\large Pangram} \\
\cmidrule(r){2-8} \cmidrule(r){9-15} \cmidrule{16-22}
\textbf{Model} & 1 & 2 & 3 & 4 & 5 & 6 & 7 & 1 & 2 & 3 & 4 & 5 & 6 & 7 & 1 & 2 & 3 & 4 & 5 & 6 & 7 \\
\cmidrule(r){1-1}\cmidrule(r){2-8}\cmidrule(r){9-15}\cmidrule{16-22}
DeepSeek-V4-Pro & \cellcolor[HTML]{FEFEFE}1 & \cellcolor[HTML]{FEFFFE}1 & \cellcolor[HTML]{FEFEFE}1 & \cellcolor[HTML]{FFFFFF}0 & \cellcolor[HTML]{FFFFFF}0 & \cellcolor[HTML]{FFFFFF}0 & \cellcolor[HTML]{FFFFFF}0 & \cellcolor[HTML]{FBFCFB}7 & \cellcolor[HTML]{FDFEFD}3 & \cellcolor[HTML]{FEFEFE}2 & \cellcolor[HTML]{FEFEFE}1 & \cellcolor[HTML]{FFFFFF}1 & \cellcolor[HTML]{FEFEFE}2 & \cellcolor[HTML]{FEFFFE}1 & \cellcolor[HTML]{FFFFFF}1 & \cellcolor[HTML]{FFFFFF}1 & \cellcolor[HTML]{FFFFFF}0 & \cellcolor[HTML]{FFFFFF}0 & \cellcolor[HTML]{FFFFFF}0 & \cellcolor[HTML]{FFFFFF}0 & \cellcolor[HTML]{FFFFFF}0 \\
Llama-3.1-8B & \cellcolor[HTML]{FEFEFE}1 & \cellcolor[HTML]{FFFFFF}0 & \cellcolor[HTML]{FFFFFF}0 & \cellcolor[HTML]{FFFFFF}0 & \cellcolor[HTML]{FFFFFF}0 & \cellcolor[HTML]{FFFFFF}0 & \cellcolor[HTML]{FFFFFF}0 & \cellcolor[HTML]{FBFCFB}6 & \cellcolor[HTML]{FEFEFE}2 & \cellcolor[HTML]{FEFEFE}2 & \cellcolor[HTML]{FEFEFE}2 & \cellcolor[HTML]{FFFFFF}1 & \cellcolor[HTML]{FEFFFE}1 & \cellcolor[HTML]{FFFFFF}1 & \cellcolor[HTML]{FBFDFB}7 & \cellcolor[HTML]{FCFDFC}5 & \cellcolor[HTML]{FDFEFD}4 & \cellcolor[HTML]{FEFEFE}2 & \cellcolor[HTML]{FEFFFE}1 & \cellcolor[HTML]{FEFEFE}2 & \cellcolor[HTML]{FEFFFE}1 \\
Qwen3.5-9B & \cellcolor[HTML]{FEFEFE}1 & \cellcolor[HTML]{FEFEFE}1 & \cellcolor[HTML]{FEFFFE}1 & \cellcolor[HTML]{FFFFFF}0 & \cellcolor[HTML]{FFFFFF}0 & \cellcolor[HTML]{FFFFFF}0 & \cellcolor[HTML]{FFFFFF}0 & \cellcolor[HTML]{FDFDFD}4 & \cellcolor[HTML]{FEFEFE}2 & \cellcolor[HTML]{FEFEFE}1 & \cellcolor[HTML]{FEFFFE}1 & \cellcolor[HTML]{FFFFFF}1 & \cellcolor[HTML]{FFFFFF}0 & \cellcolor[HTML]{FFFFFF}0 & \cellcolor[HTML]{FAFCFA}9 & \cellcolor[HTML]{FDFEFD}4 & \cellcolor[HTML]{FEFEFE}2 & \cellcolor[HTML]{FFFFFF}1 & \cellcolor[HTML]{FFFFFF}1 & \cellcolor[HTML]{FFFFFF}0 & \cellcolor[HTML]{FFFFFF}0 \\
GPT-5.6-Luna & \cellcolor[HTML]{FDFEFD}2 & \cellcolor[HTML]{FEFEFE}1 & \cellcolor[HTML]{FFFFFF}0 & \cellcolor[HTML]{FFFFFF}0 & \cellcolor[HTML]{FFFFFF}0 & \cellcolor[HTML]{FFFFFF}0 & \cellcolor[HTML]{FFFFFF}0 & \cellcolor[HTML]{FBFDFC}6 & \cellcolor[HTML]{FDFEFD}3 & \cellcolor[HTML]{FEFEFE}2 & \cellcolor[HTML]{FEFEFE}1 & \cellcolor[HTML]{FFFFFF}1 & \cellcolor[HTML]{FEFEFE}2 & \cellcolor[HTML]{FEFEFE}1 & \cellcolor[HTML]{F4F8F4}20 & \cellcolor[HTML]{FCFDFC}6 & \cellcolor[HTML]{FEFEFE}2 & \cellcolor[HTML]{FFFFFF}1 & \cellcolor[HTML]{FFFFFF}1 & \cellcolor[HTML]{FFFFFF}0 & \cellcolor[HTML]{FFFFFF}0 \\
Kimi-K2.6 & \cellcolor[HTML]{FDFEFD}2 & \cellcolor[HTML]{FEFEFE}1 & \cellcolor[HTML]{FFFFFF}0 & \cellcolor[HTML]{FEFFFE}1 & \cellcolor[HTML]{FFFFFF}0 & \cellcolor[HTML]{FFFFFF}0 & \cellcolor[HTML]{FFFFFF}0 & \cellcolor[HTML]{FCFDFC}5 & \cellcolor[HTML]{FDFEFD}4 & \cellcolor[HTML]{FEFFFE}1 & \cellcolor[HTML]{FFFFFF}1 & \cellcolor[HTML]{FFFFFF}0 & \cellcolor[HTML]{FFFFFF}0 & \cellcolor[HTML]{FFFFFF}0 & \cellcolor[HTML]{F4F8F5}20 & \cellcolor[HTML]{FAFCFA}9 & \cellcolor[HTML]{FEFEFE}2 & \cellcolor[HTML]{FFFFFF}1 & \cellcolor[HTML]{FFFFFF}0 & \cellcolor[HTML]{FFFFFF}1 & \cellcolor[HTML]{FFFFFF}0 \\
GPT-5.6-Sol & \cellcolor[HTML]{FEFEFE}1 & \cellcolor[HTML]{FEFFFE}1 & \cellcolor[HTML]{FFFFFF}0 & \cellcolor[HTML]{FFFFFF}0 & \cellcolor[HTML]{FFFFFF}0 & \cellcolor[HTML]{FFFFFF}0 & \cellcolor[HTML]{FFFFFF}0 & \cellcolor[HTML]{FAFCFA}8 & \cellcolor[HTML]{FBFCFB}7 & \cellcolor[HTML]{FCFDFD}4 & \cellcolor[HTML]{FDFEFD}3 & \cellcolor[HTML]{FEFEFE}2 & \cellcolor[HTML]{FEFEFE}2 & \cellcolor[HTML]{FEFEFE}2 & \cellcolor[HTML]{EDF4EE}32 & \cellcolor[HTML]{F6F9F7}16 & \cellcolor[HTML]{FAFCFB}8 & \cellcolor[HTML]{FEFEFE}3 & \cellcolor[HTML]{FFFFFF}1 & \cellcolor[HTML]{FEFEFE}2 & \cellcolor[HTML]{FFFFFF}1 \\
\cmidrule(r){1-1}\cmidrule(r){2-8}\cmidrule(r){9-15}\cmidrule{16-22}
HumanLM-Opinion & \cellcolor[HTML]{FFFFFF}0 & \cellcolor[HTML]{FFFFFF}0 & \cellcolor[HTML]{FFFFFF}0 & \cellcolor[HTML]{FFFFFF}0 & \cellcolor[HTML]{FFFFFF}0 & \cellcolor[HTML]{FFFFFF}0 & \cellcolor[HTML]{FFFFFF}0 & \cellcolor[HTML]{FDFDFD}4 & \cellcolor[HTML]{FDFEFD}3 & \cellcolor[HTML]{FEFEFE}2 & \cellcolor[HTML]{FEFEFE}2 & \cellcolor[HTML]{FEFEFE}2 & \cellcolor[HTML]{FFFFFF}1 & \cellcolor[HTML]{FFFFFF}1 & \cellcolor[HTML]{FFFFFF}0 & \cellcolor[HTML]{FFFFFF}0 & \cellcolor[HTML]{FFFFFF}0 & \cellcolor[HTML]{FFFFFF}0 & \cellcolor[HTML]{FFFFFF}0 & \cellcolor[HTML]{FFFFFF}0 & \cellcolor[HTML]{FFFFFF}0 \\
USP & \cellcolor[HTML]{FEFEFE}1 & \cellcolor[HTML]{FBFCFB}4 & \cellcolor[HTML]{FCFDFC}3 & \cellcolor[HTML]{FAFCFA}5 & \cellcolor[HTML]{FAFCFA}5 & \cellcolor[HTML]{F9FBF9}6 & \cellcolor[HTML]{F7FAF7}7 & \cellcolor[HTML]{FAFCFA}8 & \cellcolor[HTML]{F8FBF9}10 & \cellcolor[HTML]{F7FAF7}13 & \cellcolor[HTML]{F8FAF8}11 & \cellcolor[HTML]{F7FAF7}12 & \cellcolor[HTML]{F7FAF7}13 & \cellcolor[HTML]{F7FAF7}13 & \cellcolor[HTML]{E0EBE1}56 & \cellcolor[HTML]{E8F0E9}42 & \cellcolor[HTML]{E6EFE7}46 & \cellcolor[HTML]{EAF1EA}39 & \cellcolor[HTML]{EDF3EE}33 & \cellcolor[HTML]{EDF3ED}33 & \cellcolor[HTML]{F3F7F3}23 \\
UserLM-8B & \cellcolor[HTML]{E5EEE6}24 & \cellcolor[HTML]{E3EDE4}25 & \cellcolor[HTML]{EEF4EE}16 & \cellcolor[HTML]{F5F9F5}9 & \cellcolor[HTML]{F8FAF8}7 & \cellcolor[HTML]{FBFCFB}4 & \cellcolor[HTML]{FCFDFD}2 & \cellcolor[HTML]{E8F1E9}35 & \cellcolor[HTML]{E8F0E9}36 & \cellcolor[HTML]{ECF3EC}30 & \cellcolor[HTML]{F1F6F2}21 & \cellcolor[HTML]{F4F8F4}17 & \cellcolor[HTML]{F7FAF7}13 & \cellcolor[HTML]{F7FAF7}12 & \cellcolor[HTML]{D9E7DB}68 & \cellcolor[HTML]{E1ECE2}55 & \cellcolor[HTML]{E7EFE8}44 & \cellcolor[HTML]{EAF2EB}38 & \cellcolor[HTML]{EFF5F0}29 & \cellcolor[HTML]{F1F6F2}24 & \cellcolor[HTML]{F2F7F3}23 \\
SFTUser-14B & \cellcolor[HTML]{ECF3EC}18 & \cellcolor[HTML]{E7F0E8}22 & \cellcolor[HTML]{F2F7F3}12 & \cellcolor[HTML]{FAFCFA}5 & \cellcolor[HTML]{F7FAF7}8 & \cellcolor[HTML]{F9FBF9}6 & \cellcolor[HTML]{FAFCFA}5 & \cellcolor[HTML]{EEF4EF}26 & \cellcolor[HTML]{ECF3ED}29 & \cellcolor[HTML]{F2F7F3}20 & \cellcolor[HTML]{F6F9F6}14 & \cellcolor[HTML]{F5F9F6}15 & \cellcolor[HTML]{F5F9F6}15 & \cellcolor[HTML]{F7FAF7}13 & \cellcolor[HTML]{D9E6DA}69 & \cellcolor[HTML]{E4EDE5}50 & \cellcolor[HTML]{EEF4EE}31 & \cellcolor[HTML]{F3F7F3}22 & \cellcolor[HTML]{F4F8F4}21 & \cellcolor[HTML]{F5F8F5}18 & \cellcolor[HTML]{F5F8F5}19 \\
\modelnamesft{} & \cellcolor[HTML]{CDDFCF}45 & \cellcolor[HTML]{D2E2D4}41 & \cellcolor[HTML]{E0EBE1}28 & \cellcolor[HTML]{E7F0E8}22 & \cellcolor[HTML]{EDF3EE}16 & \cellcolor[HTML]{F1F6F2}12 & \cellcolor[HTML]{F4F8F4}10 & \cellcolor[HTML]{CDDFCF}77 & \cellcolor[HTML]{D1E2D3}70 & \cellcolor[HTML]{DBE8DC}56 & \cellcolor[HTML]{E0EBE1}48 & \cellcolor[HTML]{E6EFE7}38 & \cellcolor[HTML]{E9F1EA}33 & \cellcolor[HTML]{ECF3ED}29 & \cellcolor[HTML]{CDDFCF}91 & \cellcolor[HTML]{D8E6DA}70 & \cellcolor[HTML]{E0EBE1}56 & \cellcolor[HTML]{E6EFE7}45 & \cellcolor[HTML]{EBF2EB}37 & \cellcolor[HTML]{EDF4EE}32 & \cellcolor[HTML]{F4F8F4}20 \\
\modelname{} & \cellcolor[HTML]{E6EFE7}23 & \cellcolor[HTML]{EDF3ED}17 & \cellcolor[HTML]{F3F8F4}10 & \cellcolor[HTML]{FAFCFA}5 & \cellcolor[HTML]{FBFDFC}3 & \cellcolor[HTML]{FAFCFA}5 & \cellcolor[HTML]{FAFBFA}5 & \cellcolor[HTML]{DCE9DE}53 & \cellcolor[HTML]{DCE9DD}54 & \cellcolor[HTML]{E4EDE5}42 & \cellcolor[HTML]{EAF2EB}32 & \cellcolor[HTML]{EEF4EE}27 & \cellcolor[HTML]{F1F6F1}22 & \cellcolor[HTML]{F3F7F4}19 & \cellcolor[HTML]{CEDFD0}89 & \cellcolor[HTML]{D6E5D7}75 & \cellcolor[HTML]{DDE9DE}61 & \cellcolor[HTML]{E1ECE2}54 & \cellcolor[HTML]{E4EEE5}49 & \cellcolor[HTML]{E7EFE8}44 & \cellcolor[HTML]{E6EFE7}45 \\
\cmidrule(r){1-1}\cmidrule(r){2-8}\cmidrule(r){9-15}\cmidrule{16-22}
Human Reference & 95 & 96 & 96 & 96 & 97 & 98 & 97 & 83 & 85 & 83 & 84 & 83 & 85 & 88 & 97 & 93 & 87 & 86 & 78 & 78 & 76 \\
\cmidrule[\heavyrulewidth](r){1-1}\cmidrule[\heavyrulewidth](r){2-8}\cmidrule[\heavyrulewidth](r){9-15}\cmidrule[\heavyrulewidth]{16-22}
\end{tabular}%
}

%% file: tables/stress_test_quality_terminated.tex
\resizebox{\linewidth}{!}{%
\begin{tabular}{@{}l/*{7}{r}/*{7}{r}@{}}
 & \multicolumn{7}{c}{\large Quality} & \multicolumn{7}{c}{\large Terminated} \\
\cmidrule(r){2-8} \cmidrule{9-15}
\textbf{Model} & 1 & 2 & 3 & 4 & 5 & 6 & 7 & 1 & 2 & 3 & 4 & 5 & 6 & 7 \\
\cmidrule(r){1-1}\cmidrule(r){2-8}\cmidrule{9-15}
DeepSeek-V4-Pro & \cellcolor[HTML]{CCDFEE}99 & \cellcolor[HTML]{CBDFEE}99 & \cellcolor[HTML]{CCE0EE}99 & \cellcolor[HTML]{CCDFEE}99 & \cellcolor[HTML]{CCE0EE}99 & \cellcolor[HTML]{CCE0EE}99 & \cellcolor[HTML]{CDE0EF}97 & \cellcolor[HTML]{F2F7FB}31 & \cellcolor[HTML]{F3F8FB}28 & \cellcolor[HTML]{F5F9FC}25 & \cellcolor[HTML]{F6FAFC}23 & \cellcolor[HTML]{F9FCFD}18 & \cellcolor[HTML]{FBFDFE}14 & \cellcolor[HTML]{FFFFFF}8 \\
Llama-3.1-8B & \cellcolor[HTML]{E1ECF5}71 & \cellcolor[HTML]{E8F1F8}61 & \cellcolor[HTML]{EAF2F8}59 & \cellcolor[HTML]{EBF3F9}57 & \cellcolor[HTML]{ECF4F9}56 & \cellcolor[HTML]{EDF4F9}55 & \cellcolor[HTML]{E9F2F8}60 & \cellcolor[HTML]{F1F6FA}33 & \cellcolor[HTML]{EEF5F9}37 & \cellcolor[HTML]{E8F1F8}47 & \cellcolor[HTML]{E6F0F7}51 & \cellcolor[HTML]{E3EEF6}56 & \cellcolor[HTML]{E2EDF6}58 & \cellcolor[HTML]{E4EEF6}55 \\
Qwen3.5-9B & \cellcolor[HTML]{D6E6F2}85 & \cellcolor[HTML]{D7E7F2}84 & \cellcolor[HTML]{DAE8F3}81 & \cellcolor[HTML]{D9E8F3}81 & \cellcolor[HTML]{DCE9F3}78 & \cellcolor[HTML]{DBE9F3}78 & \cellcolor[HTML]{DEEBF4}75 & \cellcolor[HTML]{CFE2EF}91 & \cellcolor[HTML]{D0E2F0}90 & \cellcolor[HTML]{CEE1EF}92 & \cellcolor[HTML]{CFE2EF}91 & \cellcolor[HTML]{CFE1EF}91 & \cellcolor[HTML]{CFE1EF}92 & \cellcolor[HTML]{CFE2EF}91 \\
GPT-5.6-Luna & \cellcolor[HTML]{CCDFEE}99 & \cellcolor[HTML]{CBDFEE}99 & \cellcolor[HTML]{CCE0EE}99 & \cellcolor[HTML]{CCE0EE}99 & \cellcolor[HTML]{CCE0EE}98 & \cellcolor[HTML]{CCE0EE}99 & \cellcolor[HTML]{CDE0EF}98 & \cellcolor[HTML]{DBE9F3}71 & \cellcolor[HTML]{DFECF5}63 & \cellcolor[HTML]{E6EFF7}52 & \cellcolor[HTML]{EAF2F8}45 & \cellcolor[HTML]{F1F6FA}33 & \cellcolor[HTML]{F4F8FB}28 & \cellcolor[HTML]{F6F9FC}24 \\
Kimi-K2.6 & \cellcolor[HTML]{CBDFEE}100 & \cellcolor[HTML]{CBDFEE}100 & \cellcolor[HTML]{CBDFEE}100 & \cellcolor[HTML]{CBDFEE}100 & \cellcolor[HTML]{CBDFEE}99 & \cellcolor[HTML]{CBDFEE}100 & \cellcolor[HTML]{CCE0EE}99 & \cellcolor[HTML]{D6E6F1}80 & \cellcolor[HTML]{D6E6F2}78 & \cellcolor[HTML]{DBE9F3}71 & \cellcolor[HTML]{DFEBF5}63 & \cellcolor[HTML]{E5EFF6}53 & \cellcolor[HTML]{E7F0F7}50 & \cellcolor[HTML]{EBF3F9}42 \\
GPT-5.6-Sol & \cellcolor[HTML]{CBDFEE}100 & \cellcolor[HTML]{CBDFEE}100 & \cellcolor[HTML]{CBDFEE}99 & \cellcolor[HTML]{CBDFEE}99 & \cellcolor[HTML]{CBDFEE}99 & \cellcolor[HTML]{CCE0EE}99 & \cellcolor[HTML]{CCDFEE}99 & \cellcolor[HTML]{DAE8F3}72 & \cellcolor[HTML]{DEEAF4}66 & \cellcolor[HTML]{E3EEF6}56 & \cellcolor[HTML]{E8F1F7}49 & \cellcolor[HTML]{EEF4F9}38 & \cellcolor[HTML]{F0F6FA}35 & \cellcolor[HTML]{F2F7FB}30 \\
\cmidrule(r){1-1}\cmidrule(r){2-8}\cmidrule{9-15}
HumanLM-Opinion & \cellcolor[HTML]{EDF4F9}55 & \cellcolor[HTML]{E9F1F8}60 & \cellcolor[HTML]{E9F1F8}61 & \cellcolor[HTML]{E7F0F7}62 & \cellcolor[HTML]{E9F2F8}60 & \cellcolor[HTML]{EAF2F8}60 & \cellcolor[HTML]{E7F0F7}62 & \cellcolor[HTML]{F7FAFC}22 & \cellcolor[HTML]{F8FBFD}20 & \cellcolor[HTML]{F7FAFC}22 & \cellcolor[HTML]{F8FBFD}20 & \cellcolor[HTML]{F9FBFD}19 & \cellcolor[HTML]{FBFCFE}16 & \cellcolor[HTML]{F8FBFD}20 \\
USP & \cellcolor[HTML]{F8FBFD}40 & \cellcolor[HTML]{F5F9FC}44 & \cellcolor[HTML]{F5F9FC}45 & \cellcolor[HTML]{F3F7FB}47 & \cellcolor[HTML]{F2F7FB}49 & \cellcolor[HTML]{F1F7FB}49 & \cellcolor[HTML]{F4F8FB}45 & \cellcolor[HTML]{E2EDF5}59 & \cellcolor[HTML]{E4EEF6}55 & \cellcolor[HTML]{E8F1F8}47 & \cellcolor[HTML]{ECF3F9}41 & \cellcolor[HTML]{F0F6FA}34 & \cellcolor[HTML]{F3F8FB}28 & \cellcolor[HTML]{F3F7FB}29 \\
UserLM-8B & \cellcolor[HTML]{DEEBF4}74 & \cellcolor[HTML]{E5EFF6}66 & \cellcolor[HTML]{EFF5FA}52 & \cellcolor[HTML]{F7FAFD}41 & \cellcolor[HTML]{FAFCFD}38 & \cellcolor[HTML]{FFFFFF}31 & \cellcolor[HTML]{FEFFFF}32 & \cellcolor[HTML]{DCEAF4}68 & \cellcolor[HTML]{E1EDF5}60 & \cellcolor[HTML]{EBF2F8}43 & \cellcolor[HTML]{F2F7FB}30 & \cellcolor[HTML]{F5F9FC}26 & \cellcolor[HTML]{F9FBFD}19 & \cellcolor[HTML]{FCFDFE}13 \\
SFTUser-14B & \cellcolor[HTML]{DCE9F3}78 & \cellcolor[HTML]{E1ECF5}71 & \cellcolor[HTML]{E9F1F8}61 & \cellcolor[HTML]{F1F6FA}49 & \cellcolor[HTML]{F4F8FB}46 & \cellcolor[HTML]{F7FAFC}42 & \cellcolor[HTML]{F8FBFD}40 & \cellcolor[HTML]{CFE1EF}91 & \cellcolor[HTML]{D4E5F1}82 & \cellcolor[HTML]{DCE9F3}69 & \cellcolor[HTML]{E1ECF5}60 & \cellcolor[HTML]{E2EDF6}58 & \cellcolor[HTML]{E6EFF7}52 & \cellcolor[HTML]{E6EFF7}52 \\
\modelnamesft{} & \cellcolor[HTML]{D3E4F1}90 & \cellcolor[HTML]{D4E4F1}88 & \cellcolor[HTML]{D9E7F2}82 & \cellcolor[HTML]{DDEAF4}76 & \cellcolor[HTML]{E3EEF6}68 & \cellcolor[HTML]{E8F1F7}61 & \cellcolor[HTML]{E8F1F7}62 & \cellcolor[HTML]{CDE1EF}94 & \cellcolor[HTML]{D0E2F0}89 & \cellcolor[HTML]{DAE8F3}73 & \cellcolor[HTML]{E2EDF6}57 & \cellcolor[HTML]{E8F1F7}48 & \cellcolor[HTML]{EFF5FA}36 & \cellcolor[HTML]{F0F6FA}33 \\
\modelname{} & \cellcolor[HTML]{CDE0EF}97 & \cellcolor[HTML]{CDE0EF}97 & \cellcolor[HTML]{D0E2F0}94 & \cellcolor[HTML]{D3E4F0}90 & \cellcolor[HTML]{D6E6F2}85 & \cellcolor[HTML]{D9E7F2}82 & \cellcolor[HTML]{DAE8F3}80 & \cellcolor[HTML]{CBDFEE}98 & \cellcolor[HTML]{CBDFEE}97 & \cellcolor[HTML]{CFE2EF}91 & \cellcolor[HTML]{D4E5F1}82 & \cellcolor[HTML]{D8E7F2}75 & \cellcolor[HTML]{DDEAF4}66 & \cellcolor[HTML]{DEEAF4}66 \\
\cmidrule(r){1-1}\cmidrule(r){2-8}\cmidrule{9-15}
Human Reference & 89 & 96 & 91 & 87 & 85 & 82 & 83 & 100 & 100 & 100 & 100 & 100 & 100 & 100 \\
\cmidrule[\heavyrulewidth](r){1-1}\cmidrule[\heavyrulewidth](r){2-8}\cmidrule[\heavyrulewidth]{9-15}
\end{tabular}%
}